\documentclass[conference]{IEEEtran}
\IEEEoverridecommandlockouts
\usepackage{cite}
\usepackage{amsmath,amssymb,amsfonts}
\usepackage{algorithmic}
\usepackage{graphicx}
\usepackage{textcomp}
\def\BibTeX{{\rm B\kern-.05em{\sc i\kern-.025em b}\kern-.08em
		T\kern-.1667em\lower.7ex\hbox{E}\kern-.125emX}}

\usepackage{balance}
\usepackage{color}
\usepackage{url}
\usepackage{booktabs} 
\usepackage[bookmarks=false]{hyperref}
\usepackage{enumitem}
\usepackage{multirow}
\usepackage{bm}
\usepackage{array}
\usepackage{xspace}
\newcommand{\latinphrase}[1]{\textit{#1}}  
\newcommand{\etal}{\latinphrase{et~al.}\xspace}

\usepackage[none]{hyphenat}
\usepackage[numbers,sort]{natbib}

\begin{document}

\title{Replacement AutoEncoder: A Privacy-Preserving Algorithm for Sensory Data Analysis}

\author{\IEEEauthorblockN{Mohammad Malekzadeh, Richard G. Clegg}
	\IEEEauthorblockA{
		Queen Mary University of London\\
		m.malekzadeh, r.clegg@qmul.ac.uk}
	\and
	\IEEEauthorblockN{Hamed Haddadi}
	\IEEEauthorblockA{
		Imperial College London\\
		h.haddadi@imperial.ac.uk}
}


\maketitle

\begin{abstract}
An increasing number of sensors on mobile, Internet of things (IoT), and wearable devices generate time-series measurements of physical activities. Though access to the sensory data is critical to the success of many beneficial applications such as health monitoring or activity recognition, a wide range of potentially sensitive information about the individuals can also be discovered through access to sensory data  and this cannot easily be protected using traditional privacy approaches. 

In this paper, we propose a privacy-preserving sensing framework for managing access to time-series data in order to provide utility while protecting individuals' privacy. We introduce \textit{Replacement AutoEncoder}, a novel algorithm which learns how to transform discriminative features of data that correspond to sensitive inferences,  into some features that have been more observed in non-sensitive inferences, to protect users' privacy. This efficiency is achieved by defining a user-customized objective function for deep autoencoders. Our replacement method will not only eliminate the possibility of recognizing sensitive inferences, it also eliminates the possibility of detecting the occurrence of them. That is the main weakness of other approaches such as filtering or randomization. We evaluate the efficacy of the algorithm with an activity recognition task in a multi-sensing environment using extensive experiments on three benchmark datasets.  We show that it can retain the recognition accuracy of state-of-the-art techniques while simultaneously preserving the privacy of sensitive information. Finally, we utilize the GANs for detecting the occurrence of replacement, after releasing data, and show that this can be done only if the adversarial network is trained on the users' original data.

\end{abstract}


\section{introduction}
\label{sec:into}

Smart devices around us are becoming more interconnected and a large variety of data are captured by these sensors on a regular basis. From smartphones, to smart watches and wearables, many of our modern electronic devices have the ability to produce data. Since detailed, personally-identifying datasets in their original form often contain sensitive information about individuals, inconspicuous data collection can lead to major personal privacy concerns.  As an example, just ambient light sensor data~\cite{spreitzer2014pin} or accelerometer readings~\cite{owusu2012accessory} are sufficient to extract sequences of entered text on smartphones and consequently to reveal users' passwords. Moreover, Apthorpe \etal~\cite{apthorpe2017smart} examine four IoT smart home devices and find that their network traffic rates can reveal potentially sensitive user interactions even when the traffic is encrypted. Therefore, trustworthy analysis of time-series sensory data without any revelation of sensitive information to third parties is a challenging task.

Recent advances in mobile and ubiquitous computing technologies have concurrently accelerated the interest in cloud services. One of the most desired applications of these services is the recognition of current user activity and actuate a response (such as displaying alerts or turning a device on/off). However, omnipresent data gathering and user tracking leads to inherent data security and privacy risks~\cite{ziegeldorf2014privacy}. The main question is how desired statistics of personal time-series can be released to an untrusted third party without compromising the individual's privacy. More precisely, we consider how to transform time-series sensor data in such a way that after release, unwanted sensitive information cannot be inferred, but features the user wishes to share are preserved.

\begin{figure}[t!]
	\centering
	\includegraphics[scale=0.3]{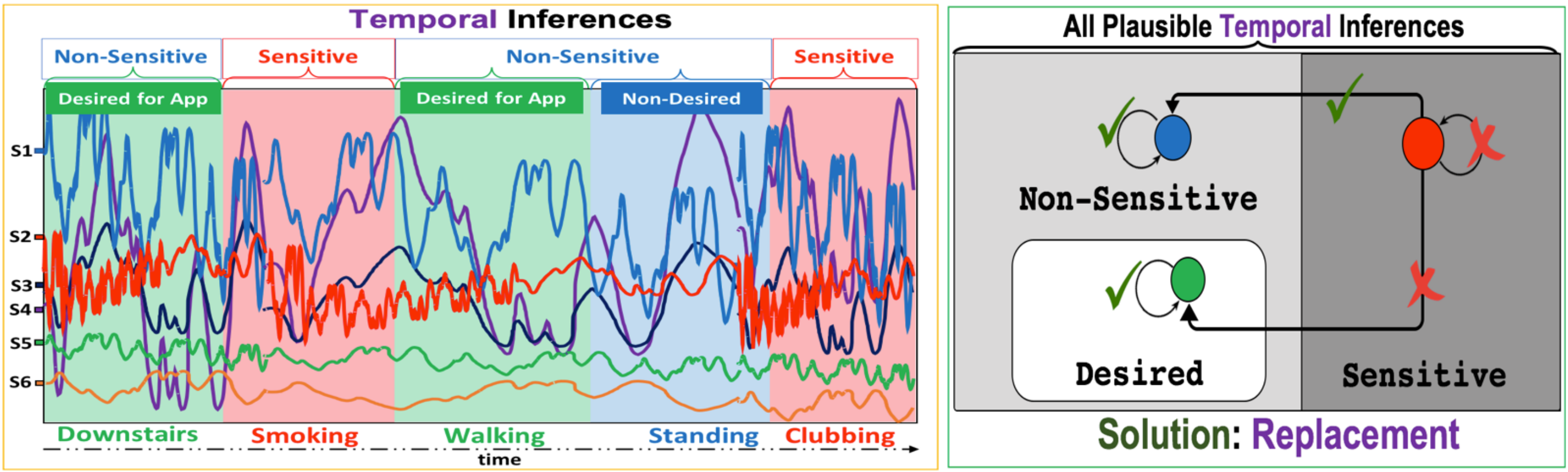}
	\caption{\label{threekind} {\it (Left)} We assume each section of a multivariate time-series contains information about a specific inference that is categorized into two classes: {\it Sensitive} and {\it Non-Sensitive} (including {\it Desired}). {(\it Right)} The ultimate goal is hiding sensitive information by transforming each sensitive section into a non-sensitive (non-desired) one while keep the remaining sections unmodified.}
\end{figure}

In this paper, thanks to recent advances in feature learning and autoencoder architectures~\cite{vincent2008extracting}, we develop a privacy-preserving platform for the analysis of multichannel time-series data generated by IoT sensors. Autoencoders are neural networks trained to reconstruct their original input, which can be considered as a form of feature extraction algorithm (we will describe it in  section~\ref{aec_def}). We consider existing approaches for preserving inference privacy in time-series data analysis and categorize the inferences that can be made  into three disjoint sets: \textit{black-listed} (sensitive for users), \textit{gray-listed} (non-sensitive for users and not desired for applications) and \textit{white-listed} (required to gain utility from third party services). Therefore, we propose a feature-based replacement method to eliminate sensitive information in time-series by transforming them to non-sensitive data  (see Figure~\ref{threekind}). By applying our method, data utility will be unaffected for specific applications and cloud apps can accurately infer the desired information, while it would be very difficult to detect and almost impossible to recognize a sensitive inference in the shared data. Specifically, our contributions are:
\begin{itemize}
	\item Inspired by denoising autoencoders~\cite{vincent2008extracting}, we introduce a deep autoencoder architecture which is able to flexibly and adaptively extract useful features from time-series data. This real-time algorithm which we call \textit{replacement autoencoder} (RAE) learns how to replace features of each section of time-series which correspond to sensitive (black-listed) inferences with some values which correspond to non-sensitive (gray-listed) inferences. 
	\item  To evaluate the tradeoff between data utility and inference privacy, we apply the algorithm to activity recognition task on three benchmark datasets collected from smart wearable devices. We show how it can retain the recognition accuracy of state-of-the-art techniques~\cite{yang2015deep,ordonez2016deep} for desired inferences, and simultaneously preserve the privacy of data subject by protecting sensitive information. 
	\item Finally, by utilizing the power of Generative Adversarial Networks (GANs)~\cite{goodfellow2014generative}, we show that adversaries can detect the presence of data replacement, by distinguishing between real gray-listed data and replaced data (transformed black-listed), only under the assumption that  they are able to access the user's  gray-listed dataset which was used to train his/her model. Though even under this assumption, recognition of the type of sensitive inferences (without any other side information) is still almost impossible.
\end{itemize}
Compared to other work on preserving the privacy of sensitive data by anonymization or randomization techniques in databases~\cite{phan2016differential, abadi2016deep}, we consider a new situation dealing with real-time sensory data. In our setting an untrusted cloud application knows the users' identity (thus differential privacy \cite{dwork2008differential} is not applicable) and we want to prevent that application inferring sensitive information from their personal data.  The proposed platform enables users to protect their sensitive data, while benefiting from sharing their non-sensitive data with cloud apps.\footnote{All the code and data used in this paper is publicly available and can be obtained from: \url{https://github.com/mmalekzadeh/replacement-autoencoder}} 

\section{Inference Privacy} \label{infpriv}
\label{sec:inference}

A time-series is a collection of observations obtained through repeated measurements over successive equally spaced points in time. The collection of sensory data, from IoT, mobile, or wearable devices, generates time-series measurements of physical quantities such as velocity and acceleration that offers tremendous opportunities for applications. For example, monitoring different physiological parameters using accelerometer and pressure sensor can create time-series for monitoring users' health status~\cite{nam2016sleep}. However, malicious programs can secretly collect the time-series and use them to discover a wide range of sensitive information about users' activities~\cite{mehrnezhad2016touchsignatures}. 

In ubiquitous sensing, the privacy of sensitive inferences is formulated as a tradeoff between users' need to receive utility from third parties with the amount of information inferred from shared time-series~\cite{Chakraborty:2013:FCP:2444776.2444791}. In this setting, it is often observed that some sections of the time-series can be used to infer \textit{sensitive} information about their users' behavior (\textit{black-listed}), and the rest of the sections contains information about \textit{desired} information (\textit{white-listed})~\cite{saleheen2016msieve}. Here, we introduce a third kind of information in sensory data: those sections of the time-series that contain \textit{non-sensitive} information (\textit{gray-listed}). Therefore, throughout this paper we discuss three kinds of inference (see Figure \ref{threekind}): 

\begin{itemize}
	\item \textbf{black-listed Inferences}: sensitive information that users wish to keep private and should not be revealed to third parties.  The black-listed inferences are sufficiently sensitive that the user would wish to prevent inference that they have undertaken any activities within this set even if the third party did not know which specific inference.
	\item \textbf{white-listed Inferences}: desired information that users gain utility from sharing with third parties services.
	\item \textbf{gray-listed Inferences}: non-sensitive information that is neutral. It is
	not useful to the user that this information is detected by third parties yet it is not important for the user to hide it. It may include data that cannot be mapped to a known behavior.
\end{itemize}

For example, consider a step counter application for a smart watch that uses time-series data generated by equipped sensors for counting the owner's steps. These rich time-series can be used for recognizing many activities, such as \textit{Walking, Jogging, Running, Sitting, Lying, Drinking, Smoking, Fighting, and Sleeping}. For an example user white-listed inferences could be \textit{Walking, Jogging, and Running}, black-listed ones could be \textit{Drinking, Smoking, Fighting, and Sleeping}, and finally, gray-listed inferences could be \textit{Sitting and Lying}. It is necessary to point out that, in comparison to static databases, there are also many possible real situations which in sensory data correspond to non-sensitive inferences. Consider the number of different possible orientations you can rotate your hand and corresponding data associated with them measured by equipped sensors in your smart watch~\cite{shen2016smartwatch}. Hence, it would be reasonable to safely replace sensitive sections of time-series with non-sensitive real sections in a way that no information about the replacement operation could leak to the other party.

 Generally, there are four privacy-preserving approaches to dealing with this kind of personal time-series: \textbf{(A)~Randomization}: adding noise to sensitive sections of the time-series to make them harder to understand. \textbf{(B)~Filtering}: removing sensitive sections of the time-series to avoid any inference. \textbf{(C)~Mapping}: projection of the original time-series into a lower dimensional space to reduce the amount of information included in data. \textbf{(D)~Replacement}: replacement of sensitive sections of the time-series with some non-sensitive data. In the following, we discuss why the first three approaches are not suitable and how we can cover all of their disadvantages by using the fourth approach.

\subsection{Randomization}
Generally, in this approach an independent~\cite{kargupta2005random} or a correlated~\cite{moon2010publishing} noise (additive or multiplicative) is added to time-series  for hiding sensitive information. Kargupta \etal~\cite{kargupta2005random} have empirically shown that independent random noise with low variance preserves very little data privacy since most of the noise can be filtered, and high variance noise completely destroys the utility of data mining. Similarly, Wang \etal~\cite{wang2017cts} argued that independent and identically distributed Laplace noise series used in current approaches can be removed from the correlated time-series by utilizing a refinement method (e.g. filtering). To remedy this problem, they proposed a correlated Laplace mechanism which guarantees that the correlation between the noise and original time-series is indistinguishable to an adversary. However, their method does not support non-stationary and high-dimensional time-series. Perturbed sections can be easily detected by adversaries and adding correlated noise to each sensitive section, based on its own structure and features, will be very challenging. The main challenge is to generate a correlated noise that after adding to data, the detection of black-listed sections of the time-series will be impossible. We show how feature-based replacement can answer this challenge.

\subsection{Filtering}
In the filtering approach, instead of perturbation or randomization, we can completely remove sensitive sections of time-series data before publishing. For example, MaskIt~\cite{gotz2012maskit} is a filtering technique that decides whether to release or suppress the current state of the smartphone users (e.g. current location). By doing this, recognition of black-listed activities will be very hard by third parties because no information about them is published. However, detection of time intervals associated with sensitive activities will be very simple, hence a side channel attack is possible. For example if an attacker gets access to user's web browser history, then they might know that currently they are in a pub, so the filtered out data could correspond to ``Drinking''. In addition, some sections of a multichannel time-series  can include both sensitive and desired information, so entirely removing them can greatly reduce the efficiency of data. We discuss how feature learning  enables us to efficiently remove sensitive information from multichannel time-series without completely filtering them out. 

\subsection{Mapping}
In this approach, instead of the original high dimensional time-series, from which inferring unnecessary sensitive information will be easy, a small set of features is extracted from data and published to third parties.  This is done in such a way that the data can be only useful for inferring desired (\mbox{white-listed}) information. In fact, feature sharing can be considered as a mapping mechanism to control and reduce the inferences possible from shared sensory data~\cite{chakraborty2012balancing}.
Laforet \etal~\cite{laforet2015individual} introduced an optimization method to generate and publish an entirely new time-series from the original ones by considering some predefined constraints on the type of information that can be inferred. Another solution would be to publish just task-relevant high-level features. Recently, Osia \etal~\cite{ossia2017hybrid}  introduced an algorithm in deep learning to manipulate the extracted features (from the primary layers of neural networks) in a way that protects the data privacy against unauthorized tasks by creating a feature set which is specific to a desired task.

The hardest part of this approach is selecting the best reduced set of features that provide a reasonable tradeoff between utility and privacy.  This approach is very dependent on the method used by third parties because we need to know if their services are compatible with this reduced set of features or not. Furthermore, it is impossible to state that a feature extracted from original data only and only contains information about white-listed information and they cannot reveal any information about black-listed activities. We explain how these challenges can be answered by applying a feature-based replacement instead of feature selection.

\subsection{Replacement} 
\label{replacement} 

As discussed in the previous sections, not only recognition of sensitive activities, but also detection of the presence of such activities in the published time-series can violate a data subject's privacy. The main idea of the replacement approach is replacing sensitive sections of time-series with sections very similar to the non-sensitive sections of the time-series. This approach benefits  the main advantages of the previously mentioned approaches while mitigating their disadvantages. 

For example, Saleheen \etal~\cite{saleheen2016msieve} argues that there is a high degree of correlation between observed human behavior and the data recorded by surrounding physiological sensors. They proposed a Dynamic Bayesian Network (DBN) model that, instead of random substitution, uses user-specific substitution as a mechanism to protect the privacy of black-listed behaviors while preserving the classification accuracy for white-listed behavior. But their model-based algorithm ignored gray-listed inferences, it is offline and assumes the availability of all time-series data before starting the replacement procedure. They also maintain a mapping database, for use in the replacement phase, that stores sensor segment of different length for each possible state. In this paper, we propose a feature-based replacement method that only needs users' personal data to provide an online and robust replacement method for privacy-preserving analysis of sensory data.

\section{Mediator Architecture} 
\label{mediator}

\begin{figure}[t!]
	\centering
	\includegraphics[scale=0.365]{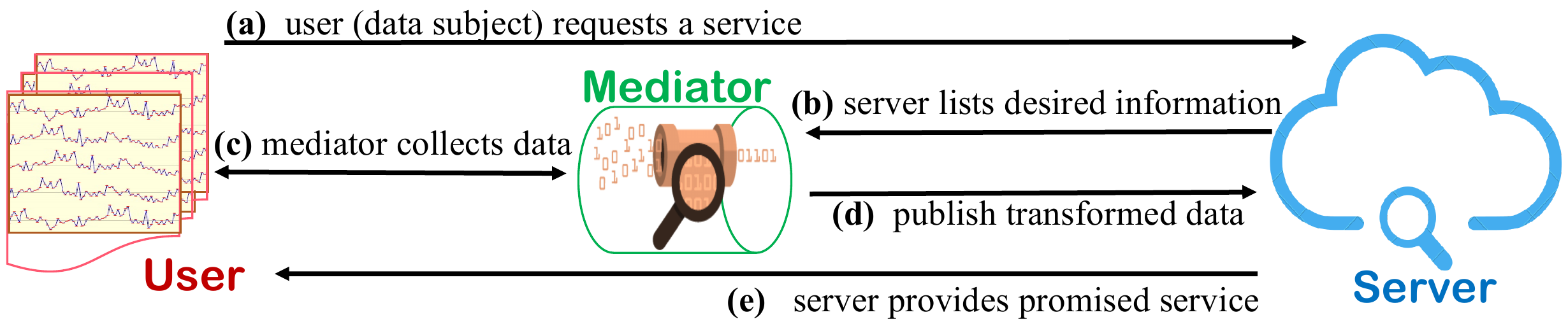}
	\caption{\label{env_arc} High-Level Architecture: a privacy-preserving mediator between cloud servers and data subject.  Sensitive sections of personal time-series are transformed with the Mediator, based on the relevant application, and then shared with the third party's server.}
\end{figure}

There are several motivations and preference factors involved when sharing personal sensor data. Users are very conservative in some situations, and less concerned in other cases. For this purpose, we need a personal and customizable framework for users to control their privacy according to their use of each application. Awareness of threats on collected time-series that can jeopardize the users' privacy will help them to choose whether to provide apps with their raw data or applying some transformation before granting access to them.  
In this section, we discuss the high-level architecture of the \textit{Mediator}: a personal platform  that enables cloud services to provide the user with desired services while protecting the privacy of the data subject (see Figure~\ref{env_arc}).

Currently, there are two major approaches to sharing sensor data for analyzing purposes. One is sharing the data with trusted third parties without any alteration based on non-disclosure agreements, but difficulties in trusting external data processors is the main weakness of this approach. Another solution is using one of the aforesaid methods in Section~\ref{infpriv}) for transforming data before publishing them to the untrusted third parties. The tradeoff between information loss and user privacy is the most challenging part of this approach. Moreover, since in the both approaches data typically leaves users' side and resides in a shared environment such as cloud servers, users no longer have any control over their data. Recently, Haddadi~\etal~\cite{haddadi2015personal} developed a personal networked device, enabling individuals to coordinate the collection of their personal data, and to selectively and transiently make those data available for specific purposes.

Considering these challenges, we propose a \textit{privacy by design} solution as an intermediate method that preserves the privacy of sensitive information without the need to trust third parties and simultaneously allow data subject to benefit from cloud services for their desired applications. Figure~\ref{env_arc} shows the high-level architecture of the Mediator; \textit{a privacy-preserving platform operates between cloud services and data subject}. Time-Series sensory data are processed with the Mediator based on the relevant application and then shared with the third party's server. In fact, based on an initial agreement between the user and the server, the Mediator intelligently hides sensitive information before releasing data. Generally, the main goal of the Mediator is to prevent the reconstruction of original time-series from transformed ones, yet allow to accurately estimate some statistics despite the transformation. 

As represented in figure~\ref{env_arc} we have three major actors here:

\begin{itemize}
	\item \textbf{Server}: a third party provides a desired service hosted in the cloud (such as activity recognition, health monitoring, occupancy detection, and smart home management). Here, we assume it owns a machine learning model that uses some of the features of the users' time-series to produce an output related to the requested service. 
	
	\item \textbf{User} : a user owns some sensors that continuously generate time-series data (such as noise, light, oxygen, temperature, acceleration, humidity, motion, and etc. ). User wants to benefit from cloud hosted services by sharing their sensory data, but in a way that only desired and non-sensitive information can be inferred.
	
	\item  \textbf{Mediator}: this is generally a machine learning platform. It creates a model for getting three major inputs: (1)~original time-series of \textit{User}, (2)~list of desired inferences for \textit{Server}, and (3)~a personal dataset of \textit{User} including sensitive and non-sensitive data to train Replacement AutoEncoder. After training, the Mediator can transforms User's data in real-time to send out to the \textit{Server}. Mediator divides time-series into a sequence of sections and releases each section separately. {\it It should definitely be a trusted application that runs on the User side}. Recently, some architectures have been proposed for mediating access to users' personal data, under their control, that can be used for this purpose~\cite{haddadi2015personal}.  
\end{itemize}
In the  following, we  propose Replacement AutoEncoder  as an efficient data transformation algorithm for deploying in the Mediator.
\section{Feature Learning} 
\label{aec_def}

In this section, we motivate and outline our proposed algorithm for online replacement of sensitive information in time-series with non-sensitive data. We discuss the core idea behind feature learning (also known as representation learning), talk about the properties of \textit{autoencoder} architecture and its usefulness for our purposes, and inspired by \textit{denoising autoencoder}, proposed by Vincent~\etal~\cite{vincent2008extracting},  introduce a new architecture, \textit{replacement autoencoder}, for real-time privacy-preserving analysis of sensory data.  We show how discriminative features extracted by replacement autoencoder  can be used to replace sensitive information in personal time-series with some non-sensitive data, without loss of utility for specific tasks. Throughout the paper, we will use the following notation:  $\bm{x}$ is an input vector (a section of time-series) and $\bm{z}$ is output vector which is generated by neural network. $\bm{b}$, $\bm{g}$, and $\bm{w}$ will respectively denote a \textit{black-listed}, \textit{gray-listed}, and \textit{white-listed} section of time-series. $\phi$ is an activation function and $L(x,z)$ is a loss function. An activation function (e.g. sigmoid function) of a node in a neural network defines the output of that node given a set of inputs, and a loss function considers the difference between predicted and true values for an instance of input data to measure how well a neural network is working.

\subsection{Autoencoder} \label{aenc_sec}
The performance of machine learning models is heavily dependent on the type of data representation and the robustness of extracted features on which they are applied. The key aspect of feature learning is the intelligent extraction of features from data and discover the informative representation needed for desired task. LeCun~\etal~\cite{lecun2015deep} argue that deep-learning frameworks are in fact representation learning methods with multiple levels of representation, obtained by composing multilayer stack of non-linear modules that each transform the representation at one level into a representation at a higher, slightly more abstract level. Bengio~\etal~\cite{bengio2013representation} have shown that the hidden layers of a multilayer neural network (starting with the raw input) can potentially learn more abstract features at higher layers of representations and reuse a subset of these features for each particular task. The earlier layers of a deep neural network contain more generic features that should be useful for many tasks, but later layers become progressively more specific to the details of the classes relevant to the desired task~\cite{bengio2009learning}. 

\begin{figure}[t!]
	\centering
	\includegraphics[width=5.8cm,height=3.9cm]{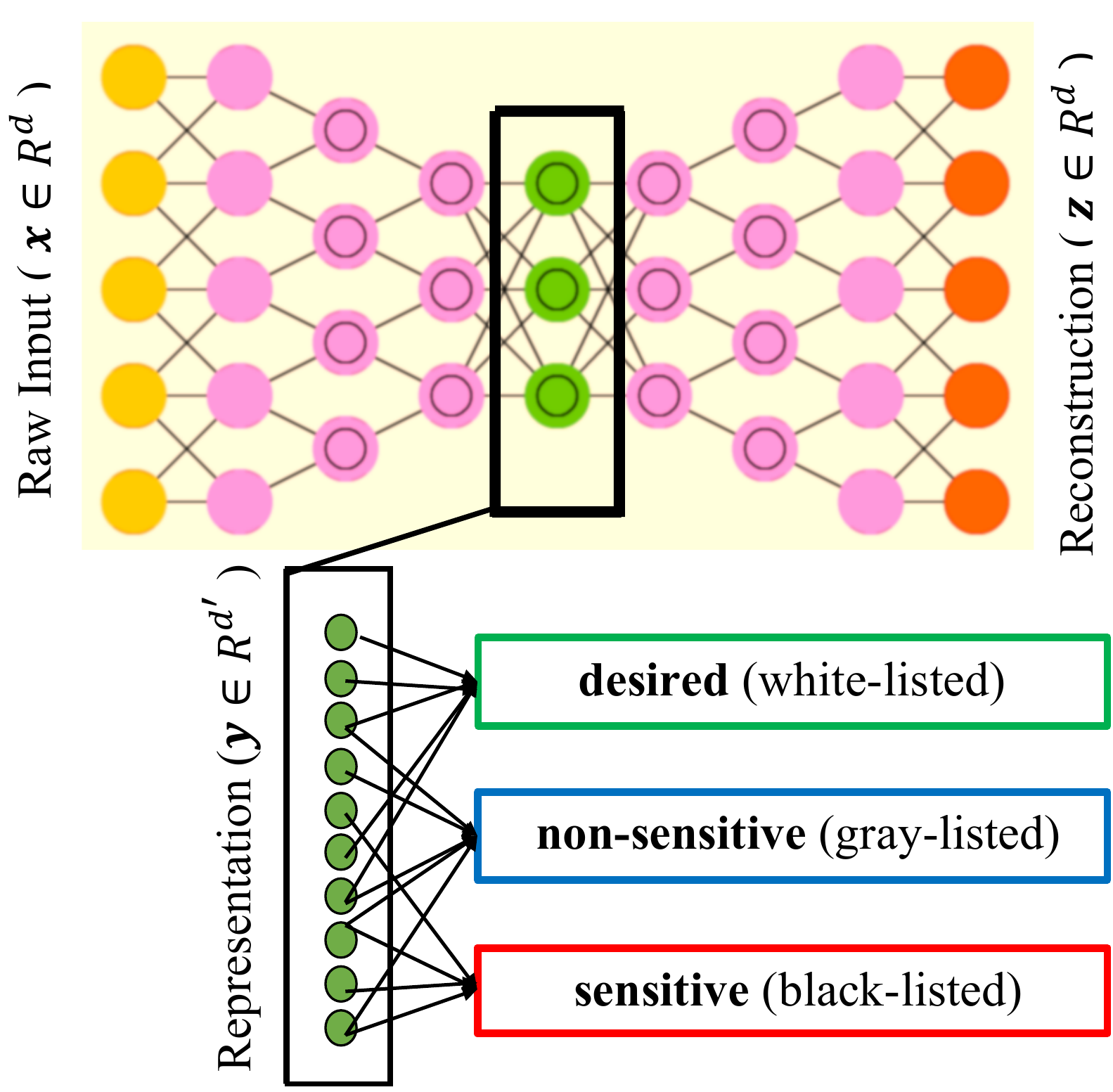}
	\caption{\label{aenc}  An autoencoder is a neural network that is trained to encode the input $x$ into some representation $y=f(x)$ so that the output $z=g(f(x))$ is a reconstruction of the input based on the representation. The autoencoder tries to capture underlying explanatory factors and discover a robust and discriminative feature set from training data. Each subset of this feature set may be relevant for each particular inference in our desired task.
	}
\end{figure} 

As depicted in Figure~\ref{aenc}, an autoencoder takes an input vector $\bm{x} =(x_{1}, x_{2}, \ldots , x_{d})  \in \mathbb{R}^d$, and first maps it to a hidden representation $\bm{y} = (y_{1},y_{2}, \ldots,y_{d{'}}) \in \mathbb{R}^{d^{'}}$ through a deterministic mapping $\bm{y} = f_{\theta}(\bm{x}) = \phi(\bm{Wx + b})$, parameterized by $\theta=\{\bm{W},\bm{b}\}$. $\phi$ is an activation function, $\bm{W}$ is a $d^{'} \times d$ weight matrix, and $\bm{b}$ is a bias vector.

Let us consider $\bm{y}$ as a feature set extracted by  autoencoder, and each $y_{i}$ in $\bm{y}$ to be equivalent to output of a node in the middlemost layer of autoencoder (see Figure~\ref{aenc}). The resulting latent representation $\bm{y}$ is then mapped back to a reconstructed vector $\bm{z} = g_{\theta^{'}}(\bm{y}) = \phi(\bm{W^{'}y + b^{'}})$ in input space ($\bm{z} \in  \mathbb{R}^d$) with $\theta^{'} = \{\bm{W^{'}},\bm{b^{'}}\}$. The parameters of the autoencoder are optimized to minimize average reconstruction error $L(\bm{x},\bm{z})$:
\begin{equation} \label{eq_aec}
\theta^* , \theta^{'*} = \arg_{\theta , \theta^{'}}\min  {{1}\over{n}} \sum_{i=1}^{n} L(\bm{x^{(i)}} , \bm{z^{(i)}})
\end{equation}
where $\bm{x^{(i)}}$ is the $i^{th}$ input data in the training dataset and $L$ is a loss function (e.g.  squared error $L(\bm{x}, \bm{z}) = ||\bm{x} -\bm{z}||^2$).
In fact, $L(\bm{x}, \bm{z})$ is a measure of the discrepancy between $\bm{x}$ and its reconstruction $\bm{z}$, over all provided training examples~\cite{vincent2010stacked}.

The autoencoder can include one or more hidden layers that hold parameters which are used to represent the input.  A deep autoencoder captures the structure of the data-generating distribution, by constraining the architecture to have a low dimension ($d^{'}<d$ in Figure~\ref{aenc}). Generally, the aim of an autoencoder is to learn a representation (encoding) for a set of data, typically for the purpose of dimensionality reduction or feature extraction. 
In an autoencoder, for each type of input (corresponding to a specific class of data), a subset of hidden units might be activated in response to it. In other words, for any given input $\bm{x}$, only a small fraction of the possible features $\bm{y^{'}} \subseteq  \bm{y} $ are relevant, thus the representation associated with that input precisely describes which features respond to the input, and how much. Deep architectures can lead to abstract representations in the last layers. In fact, more abstract concepts in the last layers can often be constructed in terms of less abstract ones captured in the early layers. More abstract concepts are generally invariant to most local changes of the input hence the outputs are usually highly nonlinear functions of the input.

\subsection{Replacement Autoencoder} \label{rep_aec_sec}

The denoising autoencoder, proposed by Vincent~\etal~\cite{vincent2008extracting}, is a special kind of autoencoder that receives a corrupted data as input and is trained to predict the original, uncorrupted data as its output. In section~\ref{aenc_sec}, we described how autoencoder minimizes a loss function $L(\bm{x}, g(f (\bm{x})))$ penalizing $g(f(\bm{x}))$ for being dissimilar from $\bm{x}$. This encourages $g(f(\bm{x}))$ to learn to be merely an identity function.  A denoising autoencoder instead minimizes $L(\bm{x}, g(f(\tilde{\bm{x}})))$, where $L$ is a loss function and $\tilde{\bm{x}}$ is a copy of $\bm{x}$ that has been corrupted by some form of noise. Denoising autoencoders must therefore undo this corruption rather than simply copying their input. 

As an extension of denoising autoencoder, we introduce \textit{Replacement AutoEncoder} (RAE): a machine learning architecture that can be utilized for transforming time-series data in a way that sensitive  information will be replaced with non-sensitive data while desired information will be retained in data. Therefore, the main objective is to train an autoencoder to produce a non-sensitive output from a sensitive input, while keeping other types of input unchanged.

\begin{figure}[t!]
	\centering
	\includegraphics[scale=0.36]{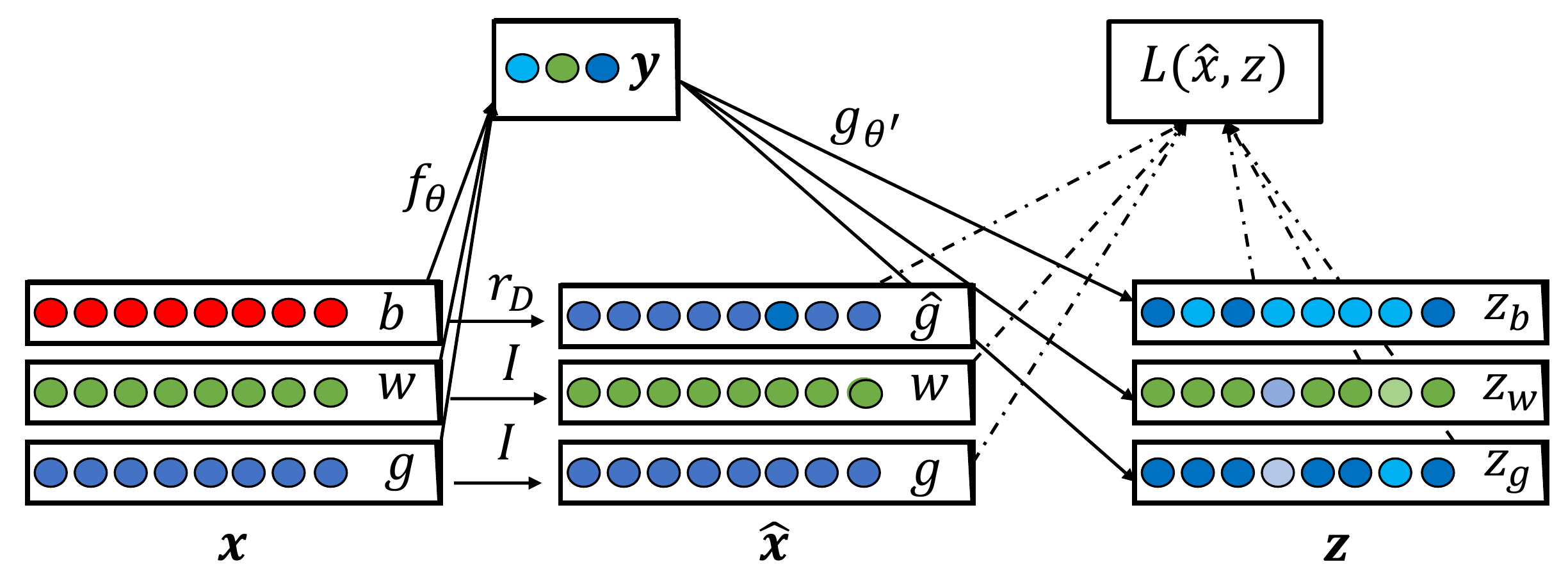}
	\caption{\label{rep_aec} Replacement Autoencoder: inspired by denoising autoencoder proposed by Vincent~\etal~\cite{vincent2008extracting}
	}
\end{figure}

As we see in the figure~\ref{rep_aec}, this is achieved by ``replacing each black-listed section $\bm{b}$ with a gray-listed section $\bm{\hat{g}}$ by means of a stochastic replacement  $ r_{D}(\bm{\hat{g}} | \bm{b})$'' in the training phase. What exactly the RAE is trying to do is implementing a piecewise function:
\begin{equation} \label{pwfunc}
\bm{z}=RAE(\bm{\hat{x}} = r_D(\bm{x})) =
\begin{cases}
\bm{x}^{SR} & \text{if $\bm{x}$ is black-listed } \\
\bm{x} & \text{otherwise} \\
\end{cases}
\end{equation}
Where we define $\bm{x}^{SR}$ as a \textit{safe replacement} for $\bm{x}$. Ideally, we say a replacement is safe if it does not reveal any information about the data replacement operation. Therefore a replaced vector $\bm{z}$ is as unsafe as the amount of information it reveals about the data replacement method.  

Here, as with the basic autoencoder, input vector $\bm{x }$ will be mapped to an intermediate discriminative representation $\bm{y = f_\theta (x) = \phi(Wx+b)}$ from which we reconstruct a $\bm{z = g_{\theta^{'}}(y) = \phi(W^{'}y + b^{'})}$. The key aspect of RAE is that in the training phase, it tries to minimize the average \textit{replacement error} $L(\bm{\hat{x}},\bm{z})$ over the training set
\begin{equation} \label{eq_raec}
\theta^* , \theta^{'*} = \arg_{\theta , \theta^{'}}\min  {{1}\over{n}} \sum_{i=1}^{n} L(\bm{\hat{x}^{(i)}} ,\bm{z^{(i)}})
\end{equation}
where, $\bm{\hat{x}}$ is a replaced version of $\bm{x}$, which can be white-listed, black-listed or gray-listed data. Again,
$\theta=\{\bm{W},\bm{b}\}$ and $\theta^{'}=\{\bm{W^{'}},\bm{b^{'}}\}$ are respectively weight matrices and bias vectors for encoding and decoding parts. RAE generates $\bm{z}$ as close as possible to the partially replaced input $\bm{\hat{x}}$. It  minimizes the criterion in equation~\ref{eq_raec} instead of equation~\ref{eq_aec}, therefore, the transformed data, $\bm{z}$, is a function of  $\bm{\hat{x}}$: the result of a stochastic replacement of $\bm{x}$ (see Figure~\ref{rep_aec}). 

The key idea of the Replacement AutoEncoder is in training phase of autoencoder: ``we train RAE in a way that it learns how to transform discriminative features that correspond to sensitive inferences, as shown in Figure~\ref{aenc},  into some features that have been more observed in non-sensitive inferences, and at the same time, keeps important features of desired inferences unchanged''.
Intuitively, RAE can be seen as an algorithm for data transformation where the related functions learned automatically from data provided rather than engineered by experts.
Note that, autoencoders only be able to reconstruct data similar to what they have been trained on. For example, an autoencoder trained on running activity data generated by an accelerometer sensor would do a rather poor job of reconstructing drinking or smoking activity data generated by the same sensor, because the features it would learn are just useful for reconstructing data correspond to running or similar activities like walking or jogging.

 In section~\ref{sec:gan} we discuss the safety of the RAE outputs. We will show that if adversaries have access to the dataset of original gray-listed data which was used to train the RAE, they can utilize it to train a Generative Adversarial Network~\cite{goodfellow2014generative} for distinguishing between a real gray-listed section of time-series and a transformed one. Therefore, although the recognition of the exact type of sensitive inferences without any side information is still impossible, personal data leak could lead to the reduced safety of replacement operation. Finally, an advantage of RAE is that it automatically learns features. For this reason, it is easy to customize them to produce a personal mediator framework, described in section~\ref{mediator}, that will perform well on a specific type of input. It doesn't require re-engineering, just appropriate personal data for training. We clarify this process  by experimental results provided in the section~\ref{exp_rslt}.

\section{Experiments} \label{expmnts}
\begin{figure*}[ht!]
		\centering
		\includegraphics[scale=0.45]{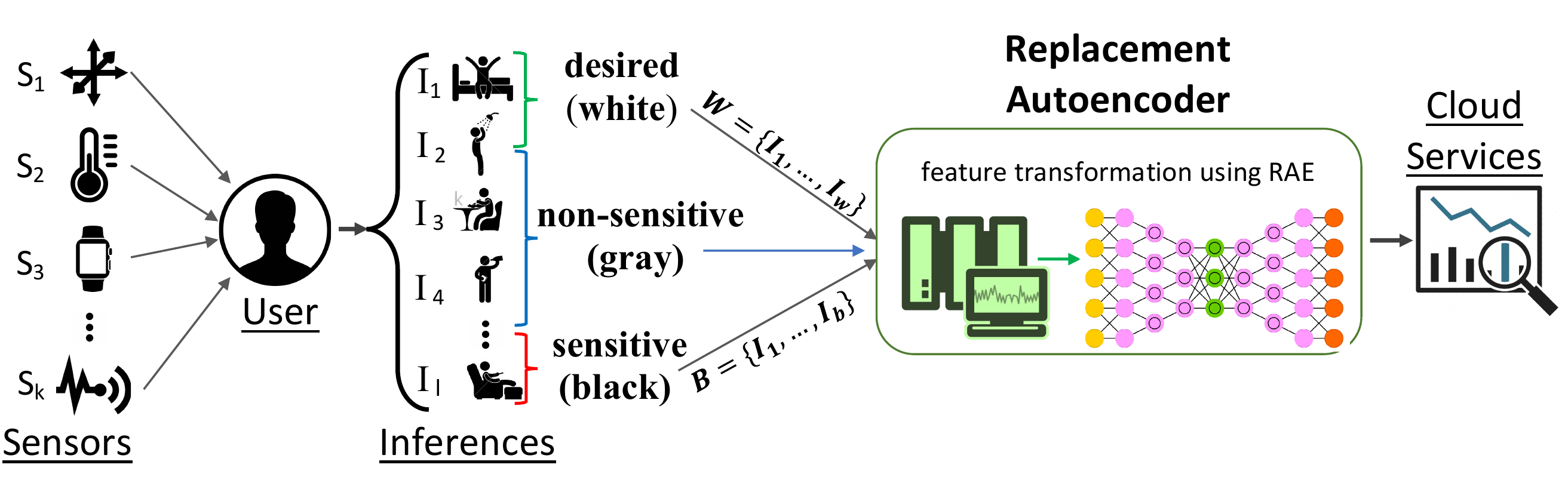}
		\caption{\label{seckeep} Replacement Platform: we assume each user knows their list of sensitive inferences and each cloud services announces their list of desired inferences. Based on defined inferences lists, RAE learns how to transform time-windows that correspond to sensitive inferences before sending them out. 
		}
\end{figure*} 
	
\subsection{Datasets} \label{dataset}
Experiments are conducted on three benchmark datasets: Opportunity~\cite{chavarriaga2013opportunity}, Skoda~\cite{zappi2008activity}, and Hand-Gesture~\cite{bulling14_csur}.

\begin{table}[t!]
	\centering
	\caption{ Activity Classes and other properties of each Dataset} \label{tbl_dataset}
	\begin{tabular}{|c|c|c|c|c|}
		\cline{1-5}
		\multirow{2}{*}{}&  \multirow{2}{*}{$\#$} &\multicolumn{3}{c|}{\textit{Name of Dataset}} \\ \cline{3-5}
		& & \textbf{Opportunity} & \textbf{Skoda} & \textbf{Hand-Gesture} \\ \cline{1-5}
		\multirow{17}{*}{\textit{Activities}}& \textbf{0} & null & null& null \\ \cline{2-5}
    	& \textbf{1} & open door1 & write notes & open window\\ \cline{2-5}
		& \textbf{2} & open door2 & open hood & close window \\ \cline{2-5}
		& \textbf{3} & close door1 & close hood & water a plant\\ \cline{2-5}
		& \textbf{4} & close door2 & check front door & turn book  \\ \cline{2-5}
		& \textbf{5} & open fridge & open left f door & drink  a bottle \\ \cline{2-5}
		& \textbf{6} & close fridge & close left f door & cut w/ knife \\ \cline{2-5}
		& \textbf{7} & open washer & close left doors & chop w/ knife \\\cline{2-5}
		& \textbf{8} & close washer & check trunk  & stir in a bowl \\ \cline{2-5}
		& \textbf{9} & open drawer1 & open/close trunk & forehand \\ \cline{2-5}
		& \textbf{10} & close drawer1 & check wheels & backhand \\ \cline{2-5}
		& \textbf{11} & open drawer2 & & smash \\ \cline{2-5}
		& \textbf{12} & close drawer2 & &\\ \cline{2-5}
		& \textbf{13} & open drawer3 & &\\ \cline{2-5}
		& \textbf{14} & close drawer3 & &\\ \cline{2-5}
		& \textbf{15} & clean table & &\\ \cline{2-5}
		& \textbf{16} & drink cup & &\\ \cline{2-5}
		& \textbf{17} & toggle switch & &\\ \cline{1-5}
		\multicolumn{2}{|c|}{\textit{Subjects}}& \textbf{4 people} & \textbf{1 person} &\textbf{2 people}\\ \cline{1-5}
		\multicolumn{2}{|c|}{\textit{Sampling Rate}}& \textbf{30 Hz} & \textbf{98 Hz} &\textbf{32 Hz}\\ \cline{1-5}
		\multicolumn{2}{|c|}{\textit{Dimension (d)}}& \textbf{113} & \textbf{60} & \textbf{15} \\ \cline{1-5}
		\end{tabular}
\end{table}

\subsubsection{Opportunity}
This dataset, famously known as the Opportunity Activity Recognition~\footnote{http://www.opportunity-project.eu}, contains naturalistic human activities recorded in a sensor environment where subjects performed daily morning activities. Two types of recording sessions were performed: Drill sessions where the subject performs sequentially a pre-defined set of activities and ``activity of daily living" runs (ADL).
Each record in the dataset comprises 113 sensory readings and the sampling rate of the sensor signals is 30 Hz~\cite{roggen2010collecting}. 
We use Drill and the first four sets of ADLs as the training data, and use the fifth set of ADL as the testing data. The data is composed of the recordings of 4 subjects and there are 18 gestures classes in this activity recognition task (Table~\ref{tbl_dataset}). The null class refers to the either non-relevant activities or non-activities. A large amount of data in this dataset matches the gray-listed \textit{Null} class (about 62\%).  For this reason it does not present much of a challenge to the RAE.  We make the task artificially harder by reducing the amount of gray-listed data by removing a high percentage of the null data. We randomly remove some instances of this class to obtain more realistic results. Therefore, we keep only 25\% of the instances of data matching the Null class to make the problem harder for RAE.

\subsubsection{Skoda}
This dataset describes the activities of assembly-line workers in a car production environment\footnote{http://www.ife.ee.ethz.ch/research/activity-recognition-datasets.html}.
They consider the recognition of 11 activity classes performed in one of the quality assurance checkpoints of the production plant (Table~\ref{tbl_dataset}). In their study, one subject wore 19 3D accelerometers on both arms and perform a set of experiments using sensors placed on the two arms of a tester (10 sensors on the \textit{right} arm and 9 sensor on the \textit{left} arm).
The original sample rate of this dataset was 98 Hz, but it was decimated to 30 Hz for comparison purposes with the other two datasets. The Skoda dataset has been employed to evaluate deep learning techniques in sensor networks~\cite{ordonez2016deep}, which makes it a proper dataset to evaluate our proposed framework.

\subsubsection{Hand-Gesture}
In this dataset, sensory data about recognizing hand gestures from body-worn accelerometers and gyroscopes is recorded\footnote{https://github.com/andreas-bulling/ActRecTut}. There are two subjects perform hand movements with 8 gestures in regular daily activity and with 3 gestures in playing tennis, so in total, there are 12 classes (Table~\ref{tbl_dataset}). Every subject repeated all activities about 26 times. Similar to the other datasets, the null class refers to the periods of no specific activity. The sampling rate of dataset is 32 Hz and each record has 15 real valued sensor readings in total. 

For all datasets, we fill up the missing values using linear interpolation and normalize each sensor's data to have zero mean and unit standard deviation. For Skoda and Hand-Gesture datasets, we randomly select 80\% of time-windows for training phase and the remaining 20\% is used for the tests.

\subsection{Experimental Settings} \label{exp_rslt}

We consider Human Activity Recognition problem (HAR) and a setting in which there are a list of $|c|$ different inference classes: $I=\{I_1, ...,I_i, ...,I_j,... , I_c\}$.
We divide inferences into three distinct categories:
\begin{enumerate}
	\item \textit{desired} activities: $I_{w}=\{I_{1}, I_{2}, ..., I_{i}\}$
	\item \textit{sensitive} activities: $I_{b}=\{I_{i+1}, I_{i+2}, ..., I_{j}\}$
	\item \textit{non-sensitive} activities: $I_{g}=\{I_{j+1}, I_{i+2}, ..., I_{c}\}$
\end{enumerate}
where $ i < j $. We assume when a service is requested from third parties, they announce the list of required inferences to be drawn to make the service works properly. We put these inferences (activities) in the list of desired inferences $I_{w}$. Moreover, we assume users know what kind of inferences are non-sensitive for them. Therefore, users announce a list of non-sensitive inferences $I_{g}$ (e.g. \textit{null} classes in datasets). In our experiments, we put the remaining inferences into sensitive inferences $I_{b}$. 
Thus, in our experiments we utilize datasets to train RAE in this way:

\begin{enumerate} [label=(\alph*)]
	\item Following~\cite{yang2015deep}, we consider a sliding window with size $d$ and step size $w$ to segment the time-series of $k$ sensors into a set of sections. Specifically, a two-dimensional matrix containing $k$ rows of time-series of which each row consists of $d$ data points. For all datasets we set $d=30$ and $w=3$. Moreover, the activity label of each sliding window is assigned by the most-frequently appeared label in all $d$ records. In this way, we have a training dataset contains labeled sections of time-series. 
	We divide training dataset into three disjoint sets, by considering $I_{b}$, $I_{w}$, and $I_{g}$ in each experiment.
	
	\item  We train the RAE for performing the \textit{feature-based replacement}. As we said in section~\ref{aec_def}, autoencoders try to learn an approximation to the identity function, so as to output $\bm{z}$ that is similar to input $\bm{x}$. Thus, for implementing the idea of Replacement AutoEncoder, we set $X = W \cup G \cup B $
	and $Z = W \cup G \cup \hat{G}$, where $X=\{x^{(1)},x^{(2)} , \ldots, x^{(n)}\}$ is input data in the training phase of $RAE$, and $Z=\{z^{(1)},z^{(2)} , \ldots, z^{(n)}\}$ is the corresponding output. $W$, $G$, and $B$ are respectively data instances corresponding to white-listed, gray-listed, and black-listed inferences.
	$\hat{G}$  is a randomly selected subset of $G$ sections with the same size with the $B$. By replacing $B$ with $\hat{G}$, $RAE$ learns how to replace sections of a time-series that contain sensitive data with non-sensitive data (implementation of equation~\ref{eq_raec} ).

	\begin{figure*}[t!]
		\centering
		\includegraphics[scale=0.5]{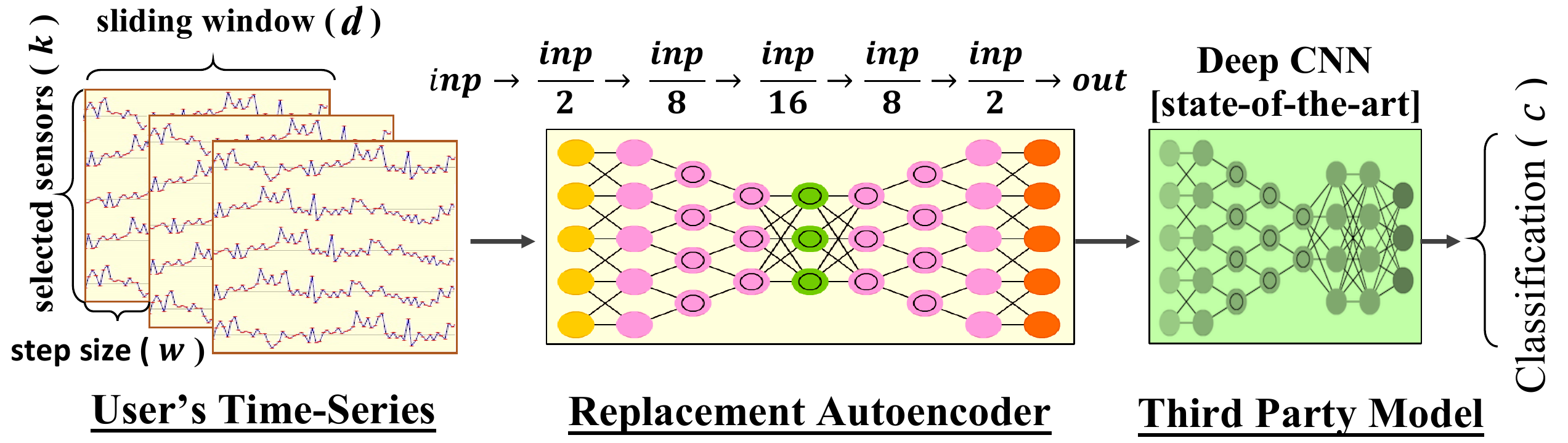}
		\caption{\label{arch}  A stacked autoencoder as a mediator : in all experiments in this paper, $w=3$ and $d = 30$ , and $k$ depends on the number of sensors in each dataset. }
	\end{figure*}

	\item We implemented RAE using Keras\footnote{https://keras.io/} framework.  The activation function for input and output layers is $linear$ and for all of the hidden layer is Scaled Exponential Linear Unit ($selu$)~\cite{klambauer2017self}. Since we normalize data to have a zero mean and unity standard deviation, $selu$ does not filter out negative numbers, thus it is a better choice than Rectified Linear Unit ($relu$). The loss function in training phase is chosen to be Mean Squared Error ($mse$), because we want to retain the overall structure of reconstructed sections.
	The dimension of RAE input is $inp=[\ k\times d, 1]\ $, and it has five hidden layers with size $inp/2 , inp/8, inp/16, inp/8, inp/2$ respectively (except for Hand-Gesture dataset, which is $inp/2 , inp/3, inp/4, inp/3, inp/2$, because of low dimensionality of data). All the experiments have been performed on 30 epochs with batch size 128. (see Figure~\ref{arch})
\end{enumerate}

\begin{table}[t!]
	\centering
	\caption{$F_1$ scores for Skoda dataset (In all the tables, $OF_1$ stands for original data and $TF_1$ stands for transformed one)} \label{cnn_accuracy}
	\begin{tabular}{|l|l|c|c|c|} 
		\hline
		Hand & List of Inferences (Table~\ref{tbl_dataset}) & $OF_1$ & $TF_1$\\ \hline
		\multirow{3}{*}{Left} &$I_{w}=\{4,8,9,10\}$ & 97.92  & 96.32 \\ \cline{2-4}
		&$I_{b}=\{1,5,6,7\}$& \textbf{96.24} & \textbf{0.00} \\ \cline{2-4}
		&$I_{g}=\{0,2,3\}$& 94.34 & 93.42 \\ \hline
		
		\multirow{3}{*}{Left} &$I_{w}=\{2,3,5,6,7,9\}$ & 96.52  & 93.23 \\ \cline{2-4}
		&$I_{b}=\{4,8,10\}$& \textbf{97.88} & \textbf{0.00}\\ \cline{2-4}
		&$I_{g}=\{0,1\}$& 93.86 & 94.85\\ \hline
		
		\multirow{3}{*}{Right} &$I_{w}=\{1,4,10\}$ & 97.56 & 94.9 \\ \cline{2-4}
		&$I_{b}=\{2,3,8,9\}$ &  \textbf{97.97} & \textbf{ 0.00} \\ \cline{2-4}
		&$I_{g}=\{0,5,6,7\}$ &  92.33 & 88.23  \\ \hline
		
		\multirow{3}{*}{Right} &$I_{w}=\{2,3,5,6,7,9\}$ & 95.76 & 91.06 \\ \cline{2-4}
		&$I_{b}=\{4,8,10\}$ &  \textbf{97.39} & \textbf{ 0.00} \\ \cline{2-4}
		&$I_{g}=\{0,1\}$ &  94.31 & 92.39  \\ \hline
	\end{tabular}

\end{table}

\begin{table}[t!]
	\centering
	\caption{$F_1$ scores for Hand-Gesture dataset } \label{cnn_accuracy2}
	
	\begin{tabular}{|c|l|c|c|c|} \hline
		Subject & List of Inferences (Table~\ref{tbl_dataset}) & $OF_1$ & $TF_1$\\ \hline
		\multirow{3}{*}{\#1} &$I_{w}=\{1,2,3,4,9,10,11\}$ & 94.11  & 90.15 \\ \cline{2-4}
		&$I_{b}=\{5,6,7,8\}$& \textbf{95.75} & \textbf{0.26} \\ \cline{2-4}
		&$I_{g}=\{0\}$& 95.04 & 96.54 \\ \hline
		
		\multirow{3}{*}{\#1} &$I_{w}=\{1,3,4,5,6,7\}$ & 95.23  & 90.45 \\ \cline{2-4}
		&$I_{b}=\{2,8,9,10,11\}$& \textbf{94.53} & \textbf{0.62}\\ \cline{2-4}
		&$I_{g}=\{0\}$&  95.04 & 97.46\\ \hline
		
		\multirow{3}{*}{\#2} &$I_{w}=\{1,3,4,5,6,7,8\}$ & 97.21& 93.30 \\ \cline{2-4}
		&$I_{b}=\{2,9,10,11\}$ &  \textbf{92.54} & \textbf{ 0.71} \\ \cline{2-4}
		&$I_{g}=\{0\}$ &  95.89 & 97.53  \\ \hline
		
		\multirow{3}{*}{\#2} &$I_{w}=\{2,3,5,6,7,9\}$ & 96.10 & 92.13 \\ \cline{2-4}
		&$I_{b}=\{4,8,10\}$ &  \textbf{96.96} & \textbf{ 0.52} \\ \cline{2-4}
		&$I_{g}=\{0,1\}$ &  95.70 & 97.56  \\ \hline
	\end{tabular}
\end{table}

\begin{table}[t!]
	\centering
	\caption{$F_1$ scores for Opportunity dataset } \label{cnn_accuracy3}
	
	\begin{tabular}{|c|l|c|c|c|} \hline
		Subject & List of Inferences (Table~\ref{tbl_dataset}) & $OF_1$ & $TF_1$\\ \hline
		\multirow{3}{*}{\#1} &$I_{w}$=\{9,10,11,12,13,14,15,16,17\} & 71.75  & 64.32 \\ \cline{2-4}
		&$I_{b}$=\{1,2,3,4,5,6,7,8\}& \textbf{79.15} & \textbf{0.21} \\ \cline{2-4}
		&$I_{g}$=\{0\}& 88.93 & 89.70 \\ \hline
		
		\multirow{3}{*}{\#1} &$I_{w}$=\{1,2,3,4,5,6,7,8,15,17\} & 76.87 & 75.93 \\ \cline{2-4}
		&$I_{b}$=\{9,10,11,12,13,14\}& \textbf{71.49} & \textbf{1.32}\\ \cline{2-4}
		&$I_{g}$=\{0,16\}&  84.44 & 82.08\\ \hline
		
		\multirow{3}{*}{\#3} &$I_{w}$=\{9,10,11,12,13,14,16\} & 74.92& 77.07 \\ \cline{2-4}
		&$I_{b}$=\{1,2,3,4,15,17\} &  \textbf{76.16} & \textbf{ 0.92} \\ \cline{2-4}
		&$I_{g}$ =\{0,5,6,7,8\}&  84.98 & 81.58  \\ \hline
		
		\multirow{3}{*}{\#3} &$I_{w}$=\{1,2,3,4,5,6,7,8,15,17\} & 70.32 & 65.05 \\ \cline{2-4}
		&$I_{b}$=\{9,10,11,12,13,14,16\} &  \textbf{74.92} & \textbf{ 6.31} \\ \cline{2-4}
		&$I_{g}$=\{0,1\} &  93.72 & 92.95  \\ \hline
	\end{tabular}
\end{table}

\begin{figure}[t!]
	\begin{minipage}[t]{0.35\linewidth}
		\centering
		\includegraphics[width=\linewidth]{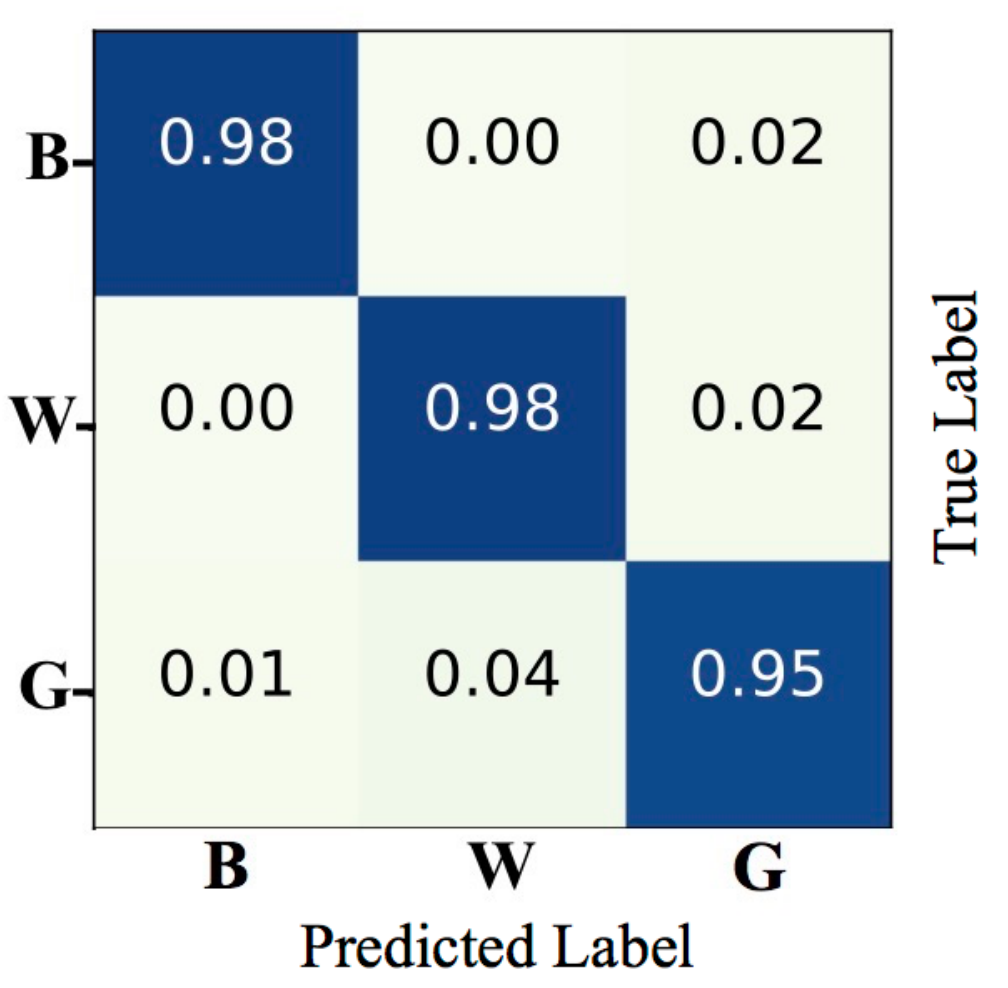}
		\label{f1}
	\end{minipage}%
	\hfill%
	\begin{minipage}[t]{0.35\linewidth}
		\includegraphics[width=\linewidth]{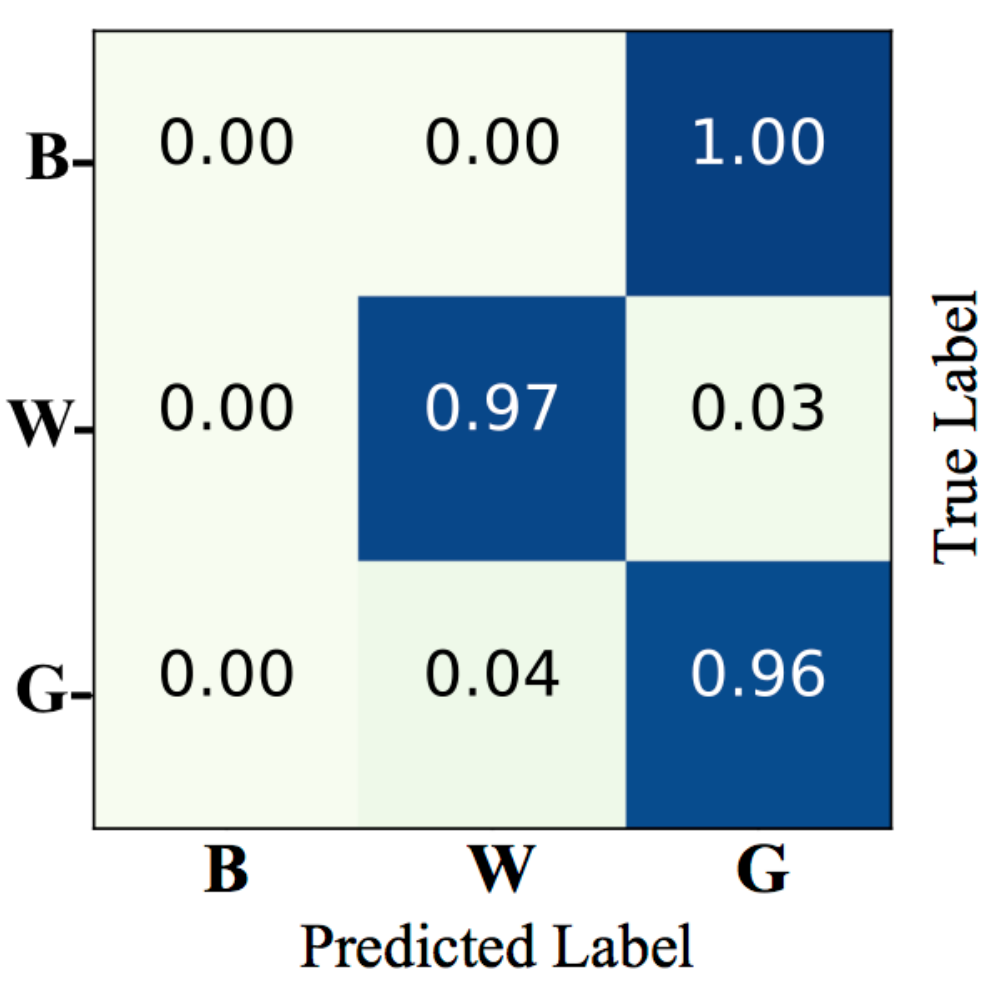}
		\label{f2}
	\end{minipage} 
	\vfill
	\begin{minipage}[t]{0.35\linewidth}
		\includegraphics[width=\linewidth]{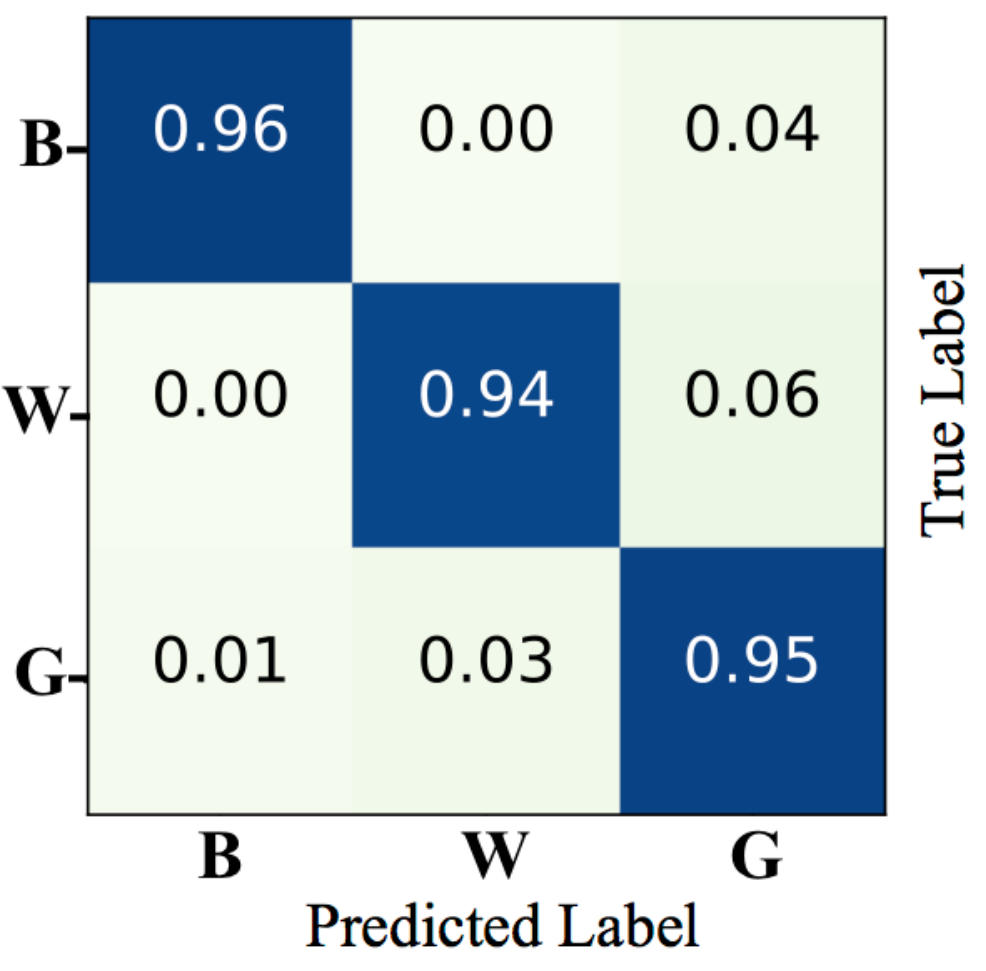}
		\label{f1}
	\end{minipage}%
	\hfill%
	\begin{minipage}[t]{0.35\linewidth}
		\includegraphics[width=\linewidth]{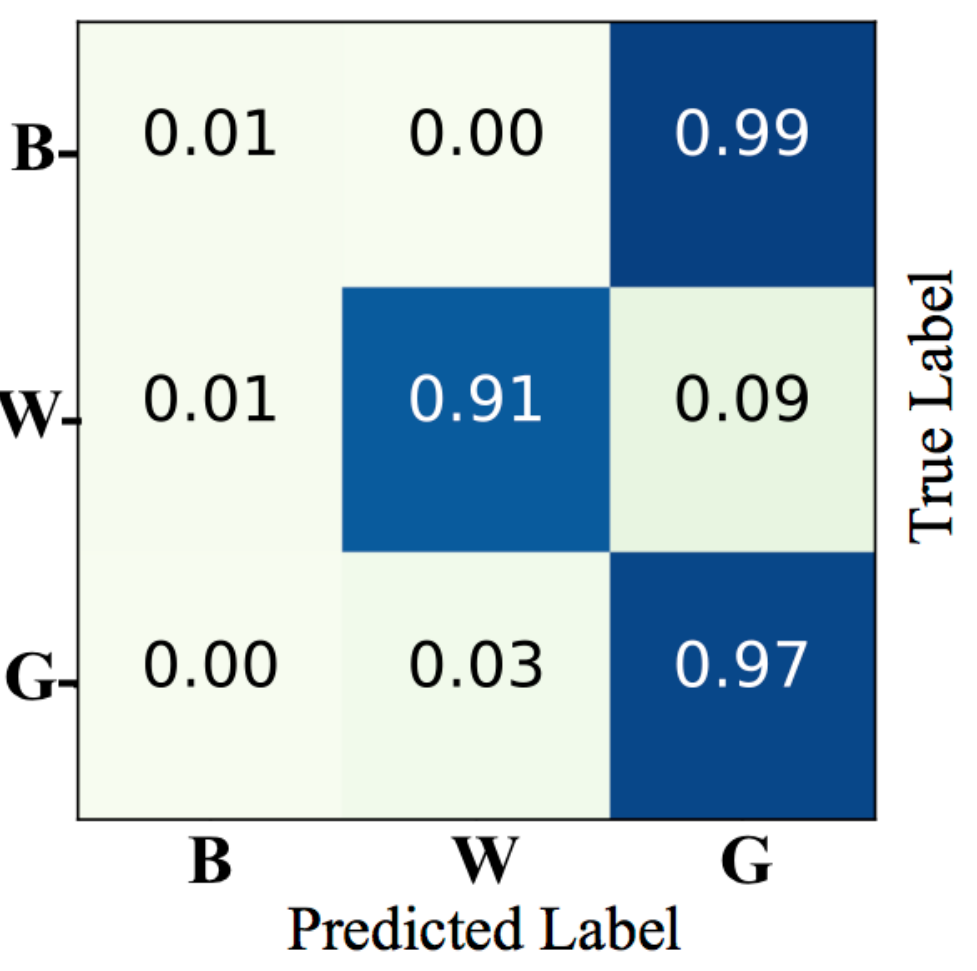}
		\label{f2}
	\end{minipage} 
	\vfill
	\begin{minipage}[t]{0.35\linewidth}
		\includegraphics[width=\linewidth]{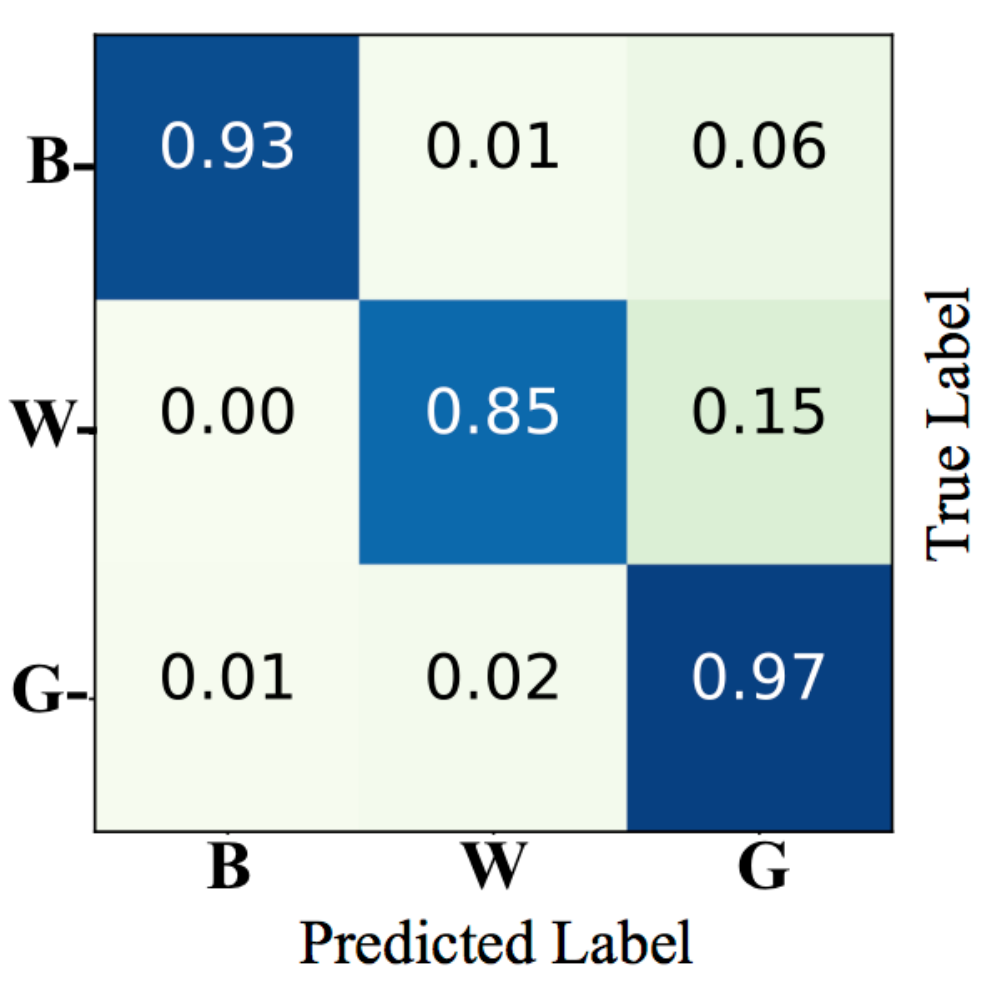}
		\label{f1}
	\end{minipage}%
	\hfill%
	\begin{minipage}[t]{0.35\linewidth}
		\includegraphics[width=\linewidth]{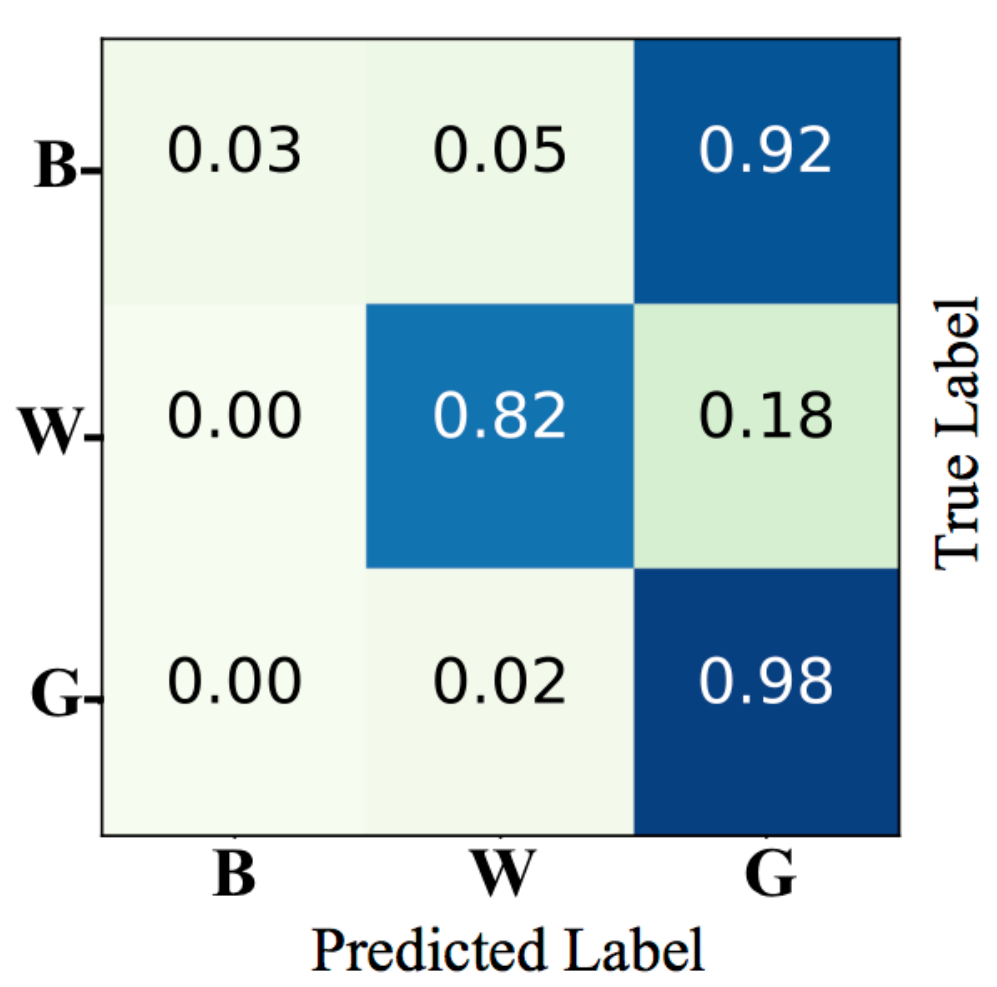}
		\label{f2}
	\end{minipage} 
	\caption{ Confusion Matrix: (Left) Original time-series . (Right) Transformed time-series. It shows that, after transformation, almost all of the black-listed activities (B) are recognized as gray-listed ones (G). (Top) Results relate to Skoda dataset (left hand) for these lists: $I_{W}=\{2,3,5,6,7,9\}$,
		$I_{B}=\{4,8,10\}$, $I_{G}=\{0,1\}$. (Middle) Results relate to Hand-Gesture dataset (subject 1) for these lists: $I_{W}=\{1,2,3,4,9,10,11\}$,
		$I_{B}=\{5,6,7,8\}$, $I_{G}=\{0\}$. (Bottom) Results relate to Opportunity dataset (subject 1) for these lists: $I_{W}=\{1,2,3,4,5,6,7,8,15,17\}$,
		$I_{B}=\{9,10,11,12,13,14\}$, $I_{G}=\{0,16\}$}\label{cnf}
\end{figure}

To evaluate the performance of time-series transformation by RAE, we implemented one of the best methods for HAR that have been proposed recently~\cite{yang2015deep} as a third party model. They build a Convolutional Neural Network to investigate the multichannel time-series data. We compare the performance of third party model on transformed time-series with original ones and show that utility will retain for all of the activities included in white-listed inferences, while recognizing an activity correspond to black-listed inferences is near impossible. Both the original and transformed time-series are given to the third party model for classification task and $F_1$ scores are calculated in Table~\ref{cnn_accuracy}, Table~\ref{cnn_accuracy2}, and Table~\ref{cnn_accuracy3}. The results show that utility will retain for desired activities, while recognizing sensitive activities is near impossible. 
The most valuable result is that third party's model misclassifies all transformed sections correspond to black-listed inferences (B) into the gray-listed inferences (G). Therefore, the false-positive rate on white-listed inferences is very low (see Figure~\ref{cnf}).

\subsection{Visualization}

It is a very challenging task to visualize multi-channel time-series, especially in our high-dimensional setting. In this section, in order to better understand the idea of Replacement AutoEncoder, we present and discuss some perceptible examples.

\begin{figure}[t!]
	\begin{minipage}[t]{\linewidth}
		\includegraphics[width=\linewidth]{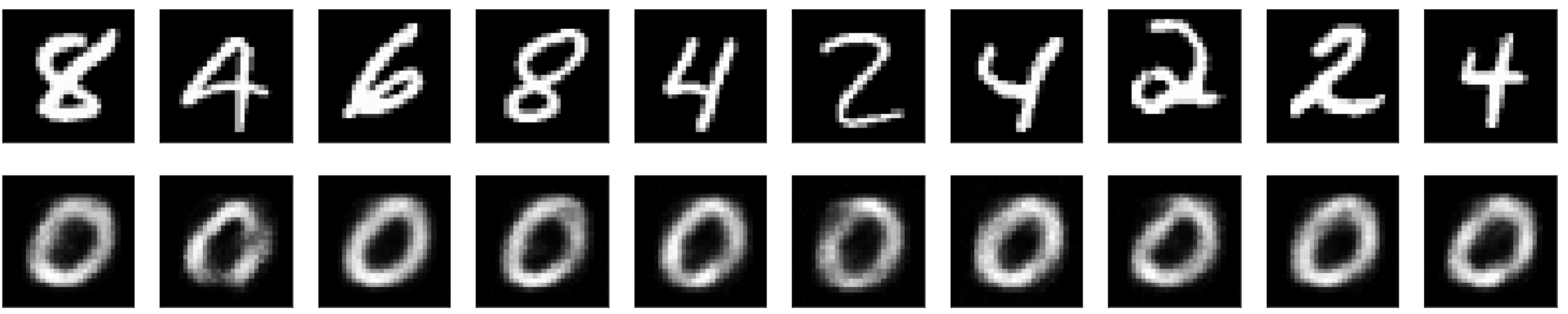}
		\label{minst1}
	\end{minipage}%
	\vfill%
	\begin{minipage}[t]{\linewidth}
		\includegraphics[width=\linewidth]{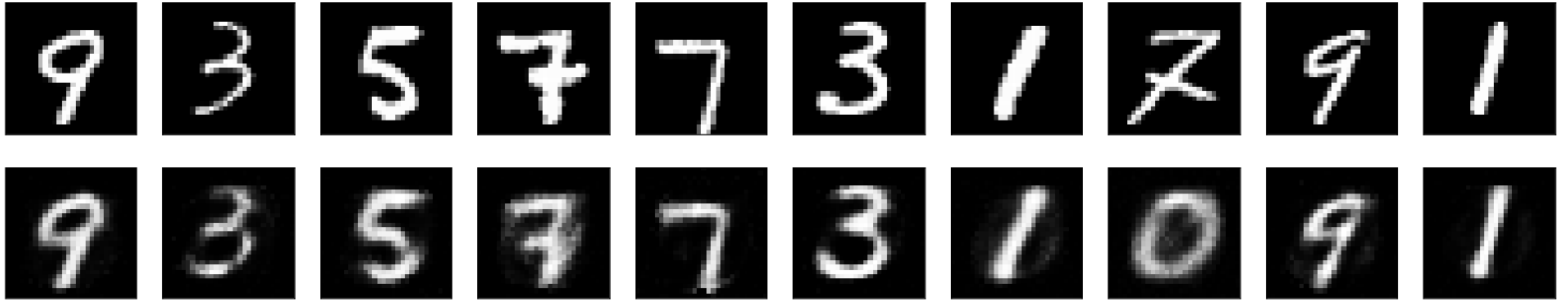}
		\label{minst2}
	\end{minipage} 
	\caption{Outputs of \textit{replacement autoencoder} in MNIST handwritten digits~\cite{lecun1998gradient}: (Top) Considering 0 as \textit{non-sensitive} (gray-listed) and information other even numbers as \textit{sensitive} (black-listed) ones. (Bottom) Considering all the odd numbers as \textit{desired} information.
	\label{mnist}}
\end{figure}

The MNIST database of handwritten digits~\cite{lecun1998gradient}, is the most famous dataset for evaluating learning techniques and pattern recognition methods. It contains $60,000$ training images and $10,000$ testing images. As an explicit example, in our setting, we consider $0$ as a kind of non-sensitive inference (gray-listed), other even numbers ($2,4,6,8$) as a set of sensitive inferences (black-listed), and all of the odd numbers ($1,3,5,7,9$) as desired information we want to keep unchanged.  Figure~\ref{mnist} shows the output of the RAE when it receives some samples from test set as input. RAE transformed all of the images categorized as sensitive information into images that are undoubtedly recognized as 0 by humans and also image recognition algorithms. This is just a simple example to build an intuitive sense of how our method can learn the piecewise function described in equation~\ref{pwfunc} to selectively censor sensitive information in personal data.

\begin{figure}[t!]
	\centering
	\begin{minipage}[t]{0.7\linewidth}
		\includegraphics[width=\linewidth]{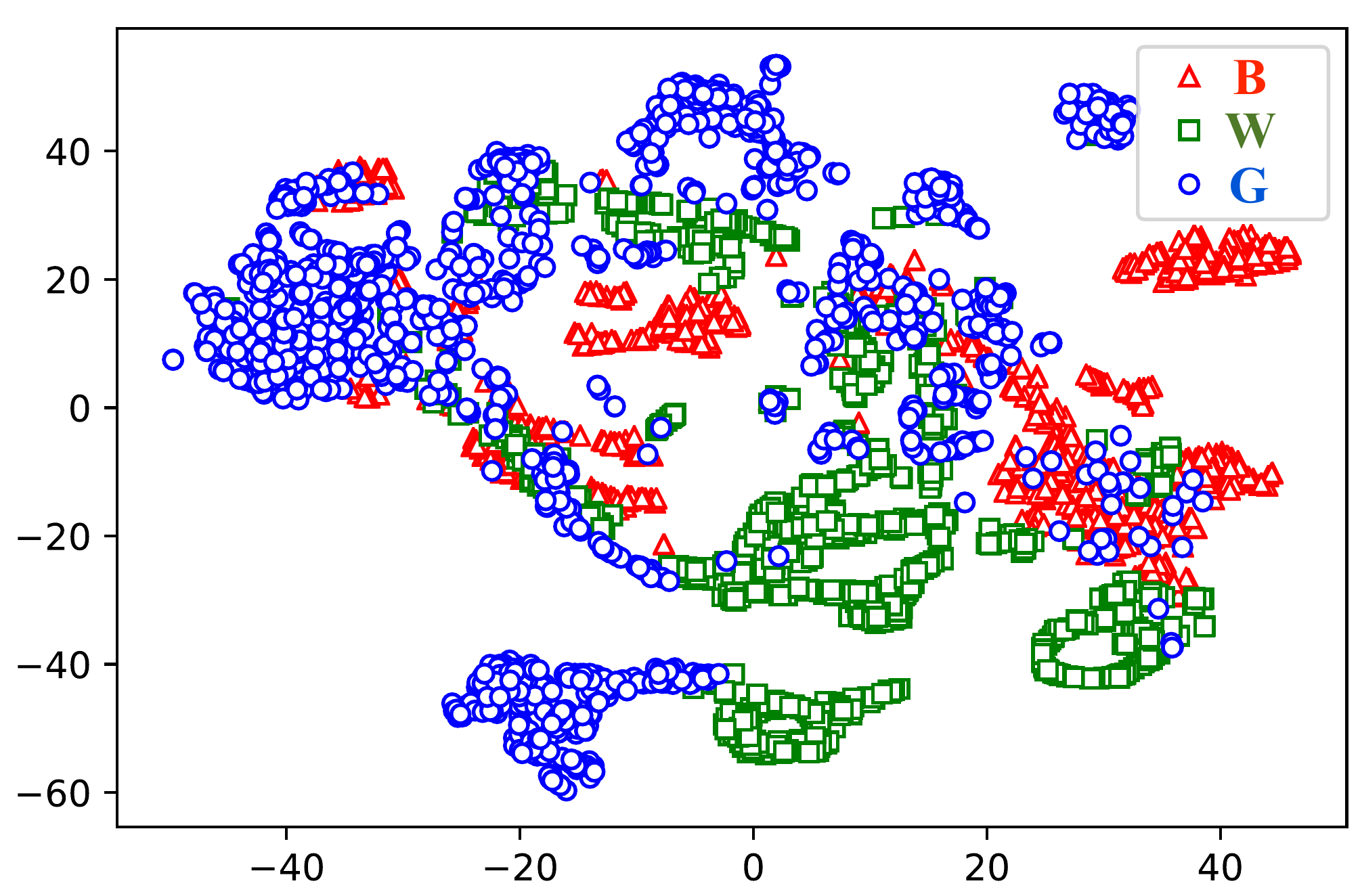}
		\label{f1}
	\end{minipage}%
	\vfill%
	\begin{minipage}[t]{0.7\linewidth}
		\includegraphics[width=\linewidth]{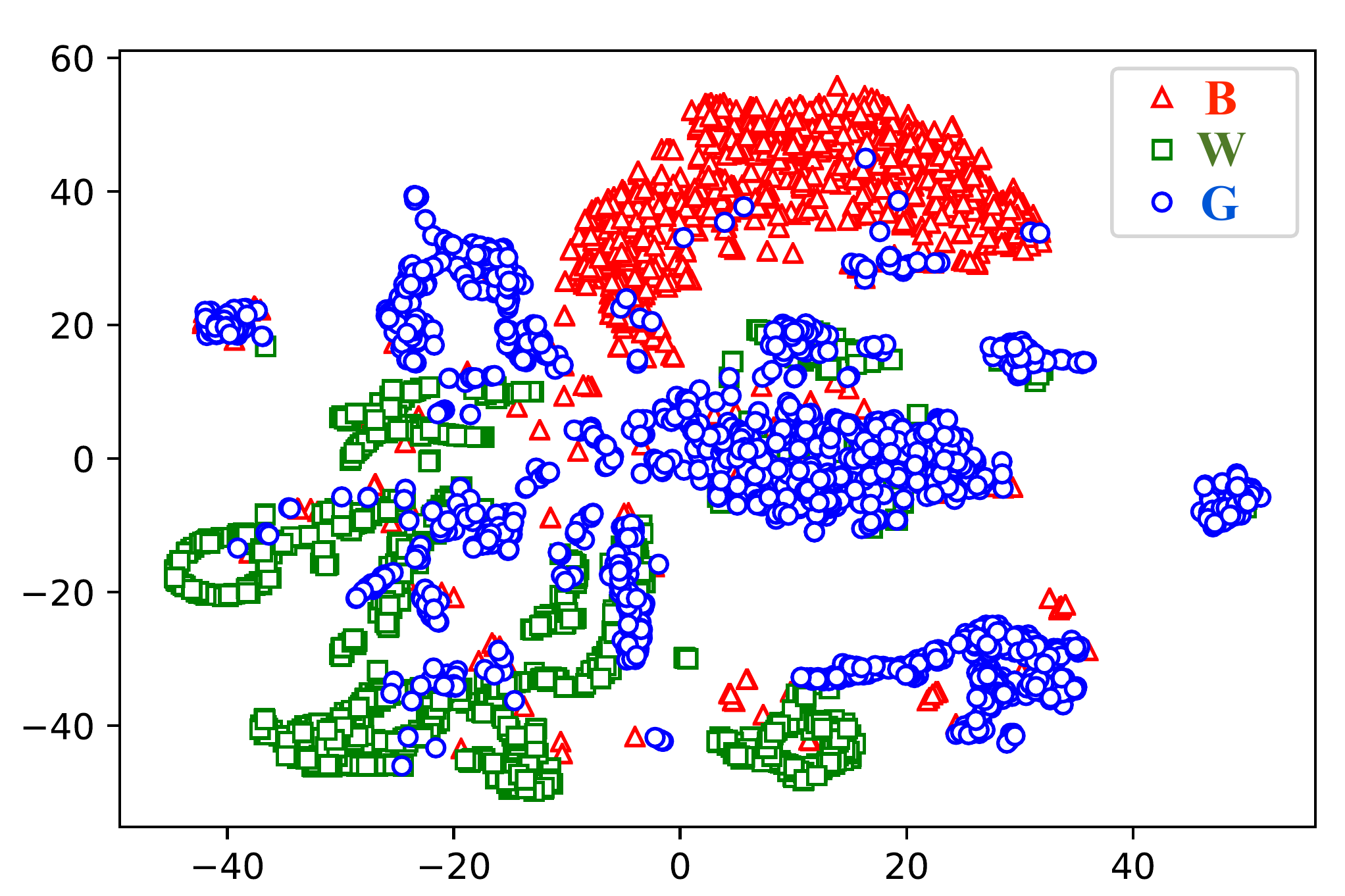}
		\label{f2}
	\end{minipage} 
	\caption{ The effect of applying feature-based transformation using replacement autoencoder on time-series: feature space relationships among different classes of activities for Skoda dataset (first row of table~\ref{cnn_accuracy} ). (Top) t-SNE on original time-series. (Bottom) t-SNE on transformed time-series.} \label{tsne}
\end{figure}

For visualizing  time-series data which are used in the experiments, we implement  t-Distributed Stochastic Neighbor Embedding (t-SNE)~\cite{maaten2008visualizing}. At the moment, it is one of the most common algorithm for 2D visualization of high-dimensional datasets.  t-SNE is a technique for dimensionality reduction and a well known strategy for visualizing similarity relationships, between different classes in the dataset, in high-dimensional data.  It uses PCA to compress high-dimensional data (e.g. 1800 for Skoda dataset~\cite{zappi2008activity}) into a low-dimensional space (e.g. 100 dimensional), then maps the compressed data to a 2D plane that can be plotted easily. Since t-SNE cares about preserving the local structure of the high-dimensional data, we use it to describe how the RAE would push sensitive data points, in the original data space, into areas that correspond to non-sensitive inferences, and at the same time preserve the separability and structures of desired data points~\footnote{ Using codes implemented in \url{https://lvdmaaten.github.io/tsne}}.

Figure~\ref{tsne} shows the feature space relationships among different classes of activities for Skoda dataset. Regarding the scatter plots depicted in Figure~\ref{tsne}, black-listed data points in original time-series (top plot) have their own structures and clearly separable from other classes. But, in the transformed version of time-series (bottom plot) all of the black-listed data points have been moved into another area of feature-space which is more closely related to gray-listed data points. They also lose their recognizable patterns and  all of them mapped to almost the same area of the feature space. On the other hand, it maintains particular patterns of white-listed data points, thus causes no harm to the recognition accuracy of desired activities. Note that in this example, we reduce the dimensionality from 1800-d to just 2-d,  which leads to huge information loss.
t-SNE has  two tunable parameters, \textit{initial dimension} (in this experiment it is equal to 100), and \textit{perplexity} (here is 60). Perplexity is a measure for information that is defined as 2 to the power of the Shannon entropy. 

\section{Threat Model} \label{sec:gan}

\begin{figure}[t!]
	\centering
	\begin{minipage}[t]{0.48\linewidth}
		\includegraphics[width=\linewidth]{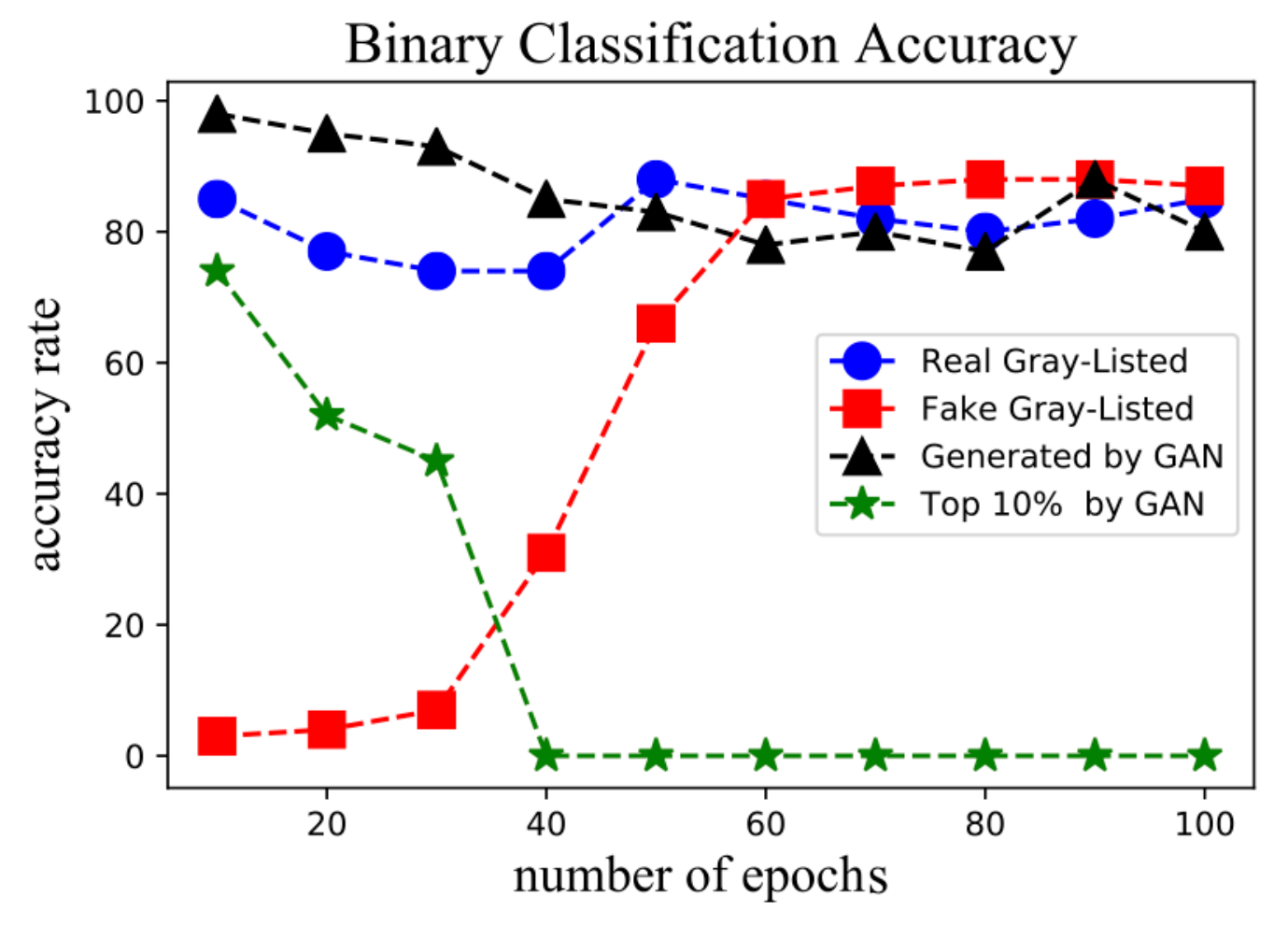}
			\caption{\label{fig:gan1} GAN's accuracy in distinguishing different kinds of real, fake and generated sections (the Skoda dataset - first row of table~\ref{cnn_accuracy}). }
		\label{f1}
	\end{minipage}%
	\hfill%
	\begin{minipage}[t]{0.48\linewidth}
		\includegraphics[width=\linewidth]{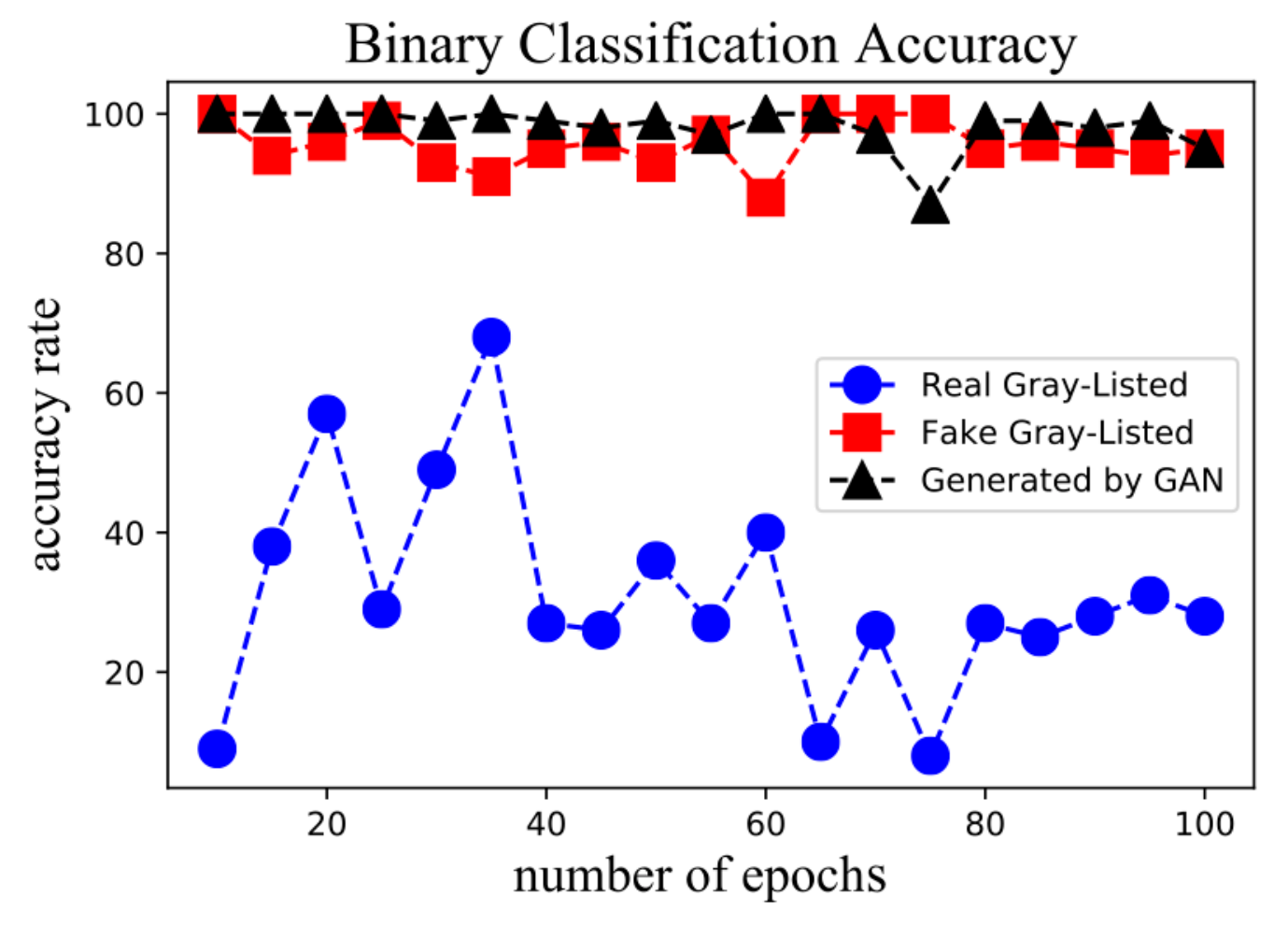}
			\caption{\label{fig:gan2}GAN's accuracy in detecting real gray-listed data by having access to other users' data (the Opportunity dataset - first row of table~\ref{cnn_accuracy3}). }
		\label{f2}
	\end{minipage} 
\end{figure}

We assume adversaries know the algorithm which is used for transforming time-series, they have access to the user's transformed time-series data, and also they have access to a dataset contain data corresponding to all the real gray-listed data which the user uses to train his/her model.  If an attacker can determine when gray-listed data from RAE is ``fake" (replacing black-listed data) then RAE is no more useful than filtering.  Hence it is important for us to show that to what extent the replacement gray-listed data that RAE writes over black-listed data is indistinguishable from the real gray-listed data output by RAE.

For this purpose, we use a Generative Adversarial Network (GAN)~\cite{goodfellow2014generative}. GANs are neural networks that learn to create synthetic data similar to some known input data. GANs consist of two models: a \textit{generative} model $G$ and a \textit{discriminative} model $D$. The model $D$ is basically a binary classifier that determines whether a given input looks like a real input from the dataset or like an artificially created data.  The model $G$ takes random input values and uses feedback from the classifier $D$ to produce convincing data that the classifier $D$ can't distinguish from real  data. Thus, we use the GAN for evaluating the output of the RAE as follows.

First, we assume that  adversaries will try to detect whether the current sections, recognized as a non-sensitive activity, are real or fake. Thus, they train a GAN on the obtained gray-listed dataset, based on the code  implemented in~\cite{radford2015unsupervised}. In this GAN, $G$ learns to produce sections very similar to real gray-listed sections and $D$ learns to determine whether the given sections, recognized as non-sensitive data, are real gray-listed sections or fake ones.
Second, after training the GAN, we separate the discriminative part $D$ and give it as input:
(i) \textit{real} gray-listed sections,
(ii)\textit{fake} gray-listed sections,
(iii)\textit{randomly generated} gray-listed sections by $G$,
and (iv)\textit{the top 10\% randomly generated} gray-listed sections as rated by discriminator $D$. We measure binary classification accuracy rate for these four categories. The results are shown in Figure~\ref{fig:gan1}.

The results demonstrate that the output of the RAE  is almost as similar as the random data generated by $G$ but not quite as good as the best data generated by $G$.  We see that when we give $D$  more time to learn, it eventually distinguishes between real gray-listed sections and fake ones. Therefore, if the user's non-sensitive data which is used to train his/her model is leaked out, it will be easy for adversaries to  distinguish between real and fake non-sensitive data. Thus, in this situation, the safety of our proposed replacement method will be reduced  to the safety of filtering approach.

We also conducted another experiment to investigate whether the privacy of user $i$ can be compromised by having access to the original gray-listed data of another user $j$. Figure~\ref{fig:gan2} shows the accuracy of the binary discriminator $D$ for distinguishing between real and fake gray-listed data of subject~\#3 when $D$ have been trained on gray-listed data from subject~\#1. Results show that $D$ cannot recognize real gray-listed data as real, and its error rate is about 70\%. Though it recognizes all fake gray-listed data as fake, it is not valuable because it recognizes real data as fake as well.
Thus, we see that a GAN can be a good discriminator only if adversaries gain access to real data from the attacked user and data from other users is not sufficient. Note that in this experiment the list of  black-listed and gray-listed inferences is the same for both data subjects.

We can see from Figure~\ref{fig:gan1} that the GAN performance in classifying data began to improve slowly from epoch 10 to epoch 30 and stabilized by epoch 70.  By contrast even after 100 epochs in Figure~\ref{fig:gan2} there is no consistent pattern of improvement for the GAN without access to the individuals gray-listed data and it seems unlikely that the GAN will ever learn to distinguish this data.

\section{Discussion and Related Work} \label{related}
 Historically, Westin~\cite{westin1968privacy} has defined \textit{information privacy} as ``the right to select what personal information about me is known to what people''. Recently, Ziegeldorf et al.~\cite{ziegeldorf2014privacy} extends this definition for \textit{IoT privacy} by ``having individual control over the processing of personal information and awareness of the subsequent use of them".  Here, we discuss the recent progress in the protection of users' privacy in time-series data analysis and compare them with our approach.
 
 \textbf{Anonymization}. In the past decade, there has been some remarkable progress in the field of preserving privacy in tabular data publishing, such as \textit{k-anonymity}~\cite{sweeney2002k} and \textit{l-diversity}~\cite{Machanavajjhala06l-diversity:privacy}. These approaches, more known as \textit{anonymization} methods, mainly protect the identity of the data owners by coarse-graining methods. They try to retain essential information for analysis but suppress or generalize other identity attributes. In this paper, instead of anonymity, we consider a situation where untrusted third parties know users' identity and we introduce a method which prevents inferring sensitive information from their personal time-series. 
 
\textbf{ Randomization}. An approach for protecting sensitive inferences is \textit{data randomization} which perturbs data by multiplicative or additive noise, or other randomized processes.
Zhang et al.~\cite{zhang2015time} developed a time-series pattern based noise generation strategy (called noise obfuscation) for privacy protection on the cloud. It generates and injects noise service requests into real ones to ensure that their occurrence probabilities are about the same so that service providers cannot distinguish which requests are real ones. To protect the user from inference attacks on his private data, Erdogdu et al.~\cite{erdogdu2015privacy} proposed a method that sequentially randomizes samples from the time-series prior to their release according to a stochastic process, called the privacy mapping. For example, a household is willing to give the aggregate electrical load to the service provider, but wishes to keep the information related to their eating patterns private, in particular the microwave usage which can also be inferred from the aggregate load. 
Allard et al.~\cite{allard2015chiaroscuro} proposed Chiaroscuro, a solution for clustering personal time-series that are distributed on personal devices with privacy guarantees. Chiaroscuro allows the participating devices to collaborate privately by combining encryption with differential privacy~\cite{dwork2008differential}.
Generally, because of high temporal and spatial granularity of time-series and strong correlation between their samples, when general data randomization approaches are extended to the dynamic case of time-series data, scalability challenges arise and maintaining utility often becomes challenging~\cite{erdogdu2015privacy}. For this reason, we focus on  a scalable method for effectively hiding sensitive information by only replacing sensitive sections of time-series without perturbing the desired and non-sensitive data.

\textbf{Deep Learning Frameworks:}
Liu et al.~\cite{liu2016collaborative} proposed a collaborative privacy-preserving deep learning system in mobile environment, which enables multiple sites to learn deep learning model only by sharing partial parameters with each other. In their approach, the privacy is protected by keeping data in local and use a parameter selection mechanism to only share small fraction of the parameters to the server at each round. 
Phan et al.~\cite{phan2016differential} enforce $\epsilon$-differential privacy  by injecting noise into the objective functions of the deep autoencoders at every layer and training step. By considering the need to protect sensitive inferences, we introduce a novel algorithm for feature learning which can be utilized for real-time privacy-preserving time-series publishing. It uses  a new objective function for learning a deep autoencoder by automatically extraction of discriminative features from input time-series .

\textbf{Discussion:}
Here we discuss some important advantages of RAE for privacy-preserving analysis of sensory data. 

\begin{itemize}
	\item \textit{No sensitive activities can be inferred}: RAE is trained to transform blacklisted sections into the same-size sections that are very similar to gray-listed activates. Therefore, on the cloud side, only non-sensitive gray-listed activities can be inferred.
	
	\item \textit{No loss in utility}: autoencoder transformed white-listed sections with the least amount of changes, because it is trained to copy these sections as exactly as possible. In addition, it will learn to transform both gray-listed and black-listed activities to gray-listed ones. So, the amount of false positive (wrongly recognizing an activity as white-listed) is near to zero. This is very important to the quality of services provided to the user.
	
	\item 	\textit{No detection of sensitive intervals}: in \textit{noise addition} and \textit{filtering}, third parties can easily detect time intervals corresponding to sensitive activities. They can use this knowledge alongside other side  information to conclude about the type of related activity. In our feature-based replacement approach it is very hard to distinguish between sensitive and non-sensitive time interval, unless adversaries get access to the original gray-listed data which have used for training RAE. 
	
	\item \textit{No hand selected feature set}: in the \textit{mapping} (feature publishing) approach we need to go through the reduced representation of a time-series and divide them into white-listed and black-listed feature sets. In feature-based replacement we publish transformed version of time-series with the same dimensionality of original data and have not concerned about separating features. In fact RAE can automatically transform features correspond to black-listed data while retaining the features correspond to white activities unchanged. 
\end{itemize} 

\section{Conclusions and Future Directions} \label{future}
The protection of users' privacy in time-series analysis is a very challenging task, especially when time-series are gathered from different sources. Invasion of privacy arises when disseminated data can be used to draw sensitive inferences about the data subject. 
In this paper, we have focused on the inference privacy and considered time-series generated by sensors embedded into mobile and wearable devices. We have introduced Replacement AutoEncoder: a feature learning algorithm which learns how to transform discriminative features that correspond to sensitive inferences,  into some features that have been more observed in non-sensitive data, and at the same time, keeps important features of desired inferences unchanged to benefit from cloud services. We evaluate the efficacy of the method with an activity recognition task using experiments on three benchmark datasets and show that it can preserve the privacy of sensitive information while simultaneously retaining the recognition accuracy of state-of-the-art techniques .

We believe that the approach of learning privacy-related features from time-series data will enable us to develop efficient methods for data transformation and help us to enrich existing IoT platform with a robust privacy-preserving time-series data analytics component. Thus, in the future, we will  focus on extending the mediator framework by letting users dynamically define their personal privacy policies and inferences that applications are allowed to access. We will also investigate theoretical frameworks and other mathematical tools to determine a bound on sensitive information that remains in the transformed time-series. Beside GANs, we will pay attention to other possible attacks and appropriate responses to them as well as comparing the achieved utility-privacy tradeoff of replacement with other privacy approaches, such as randomization or filtering.
Another direction for continuing this research could be using Long Short Term Memory (LSTMs) networks~\cite{hochreiter1997long} which are capable of learning long-term dependencies in data to capture the discriminative features of time-series.

\section*{Acknowledgment}
This work was kindly supported by the Life Sciences Initiative at Queen Mary University London and a Microsoft Azure for Research Award. Hamed Haddadi was partially funded by EPSRC Databox grant (Ref: EP/N028260/1).

\footnotesize

\end{document}